\documentclass{article}

\usepackage{microtype}
\usepackage{graphicx}
\usepackage{subfigure}
\usepackage{booktabs} 

\usepackage{hyperref}

\usepackage[accepted]{icml2023}

\usepackage{amsmath}
\usepackage{amssymb}
\usepackage{mathtools}
\usepackage{amsthm}

\usepackage[capitalize,noabbrev]{cleveref}

\theoremstyle{plain}

\theoremstyle{definition}

\theoremstyle{remark}

\usepackage[textsize=tiny]{todonotes}

\icmltitlerunning{Beyond Hawkes: Neural Multi-event Forecasting on Spatio-temporal Point Processes}

\begin{document}

\twocolumn[
\icmltitle{Beyond Hawkes: Neural Multi-event Forecasting on \\Spatio-temporal Point Processes}



\icmlsetsymbol{equal}{*}

\begin{icmlauthorlist}
\icmlauthor{Negar Erfanian}{yyy}
\icmlauthor{Santiago Segarra}{yyy}
\icmlauthor{Maarten de. Hoop}{yyy}
\end{icmlauthorlist}

\icmlaffiliation{yyy}{Department of Electrical and Computer Engineering, Rice University, Houston, Texas, USA}

\icmlcorrespondingauthor{Negar Erfanian}{ne121@rice.edu}

\icmlkeywords{Machine Learning, ICML}

\vskip 0.3in
]



\printAffiliationsAndNotice{}  

\begin{abstract}
Predicting discrete events in time and space has many scientific applications, such as predicting hazardous earthquakes and outbreaks of infectious diseases. History-dependent spatio-temporal Hawkes processes are often used to mathematically model these point events. However, previous approaches have faced numerous challenges, particularly when attempting to forecast one or multiple future events. In this work, we propose a new neural architecture for simultaneous multi-event forecasting of spatio-temporal point processes, utilizing transformers, augmented with normalizing flows and probabilistic layers. Our network makes batched predictions of complex history-dependent spatio-temporal distributions of future discrete events, achieving state-of-the-art performance on a variety of benchmark datasets including the South California Earthquakes, Citibike, Covid-19, and Hawkes synthetic pinwheel datasets. More generally, we illustrate how our network can be applied to any dataset of discrete events with associated markers, even when no underlying physics is known.
\end{abstract}

\section{Introduction}
\label{sec: intro}
Predicting the occurrence of discrete events in time and space has been the focus of many scientific studies and applications. Problems such as predicting earthquake hazards \citep{ogata1998space, chen2020neural}, infectious diseases over a population \citep{meyer2012space, schoenberg2019recursive}, mobility and traffic in cities \citep{du2016recurrent}, and brain neuronal spikes \citep{perkel1967neuronal} fall under this category and have gained quite some interest. 
Over the years, many works used the history dependent spatio-temporal Hawkes process \citep{ozaki1979maximum, ogata1988statistical, ogata1998space, ogata2006space, ogata1993fast, nandan2017objective, sornette2005constraints, zhuang2012long, helmstetter2003predictability, bansal2013non} to model these point events. The stochasticity and the excitatory history-dependency of the Hawkes process are modeled by a conditional intensity function that varies in time and space. Assuming $n$ events and their associated markers $(t, {\mathbf{x}},M)$ collected in the history $H_t= \{(t_i, {\mathbf{x}}_i,M_i) \, | \, t_i<t, i=1:n\}$, the Hawkes intensity function is defined as 
\begin{align}
\label{eq:lambda}
    \lambda(t,{\mathbf{x}}|H_t)
    &:=\lim_{\Delta t,\Delta {\mathbf{x}}\to 0}\frac{P_{\Delta t,\Delta {\mathbf{x}}}(t,{\mathbf{x}}|H_t)}{|B({\mathbf{x}},\Delta {\mathbf{x}})|\Delta t}\\ \nonumber
    &= \mu_{\theta}({\mathbf{x}}) + \sum_{i:t_i<t} \, g_{\phi}(t-t_i, {\mathbf{x}}-{\mathbf{x}}_i,M_i),
\end{align}
where $P_{\Delta t,\Delta {\mathbf{x}}}(t,{\mathbf{x}}|H_t)$ denotes the history-dependant probability of having an event in a small time interval $[t, t+\Delta t)$ and a small ball $B({\mathbf{x}},\Delta {\mathbf{x}})$ centered at ${\mathbf{x}} \in \mathbb{R}^d$ with the radius of $\Delta {\mathbf{x}}\in \mathbb{R}^d$, where $d$ denotes the dimension in space, and 
$H_t$ represents all events with their associated markers happening up to but \emph{not} including time $t$. The functions $\mu_{\theta}$ and $g_{\phi}$ represent the (parametric) background intensity function and spatio-temporal kernel, respectively, forming $\lambda(t,{\mathbf{x}}|H_t)$. A variety of different parametric forms for both $\mu_{\theta}$ and $g_{\phi}$ have been proposed~\citep{ogata1998space}. Despite their use across different domains, Hawkes processes have several limitations. 
\textbf{\textit{First}}, one has to predetermine parametric forms for $\mu_{\theta}$ and $g_\phi$. This limits the expressive power of the model and, more importantly, might lead to a mismatch between the proposed model and one that can better represent the observed data. 
\textbf{\textit{Second}}, (\ref{eq:lambda}) imposes similar behavior across all events over time meaning that every preceding event has the same form of impact on future events' occurrence. As an example, in the classical Hawkes process with a temporal decaying kernel $g_{\phi}(t-t_i) = \frac{1}{\beta}\exp(-\frac{(t-t_i)}{\beta})$, the same parameter $\beta$ is applied to all events via their corresponding time differences.
\textbf{\textit{Third}}, the intensity function $\lambda(t,{\mathbf{x}}|H_t)$ can be used to predict only one step ahead via, e.g., the commonly-used first-order moment \citep{rasmussen2011temporal, snyder2012random} given by
\begin{align}
\label{eq: firstmoment}
    \mathbb{E}[t_{n+1}|H_{t_n}] = &\int_{t_n}^\infty t \int_{\boldsymbol{V}} \lambda(t,{\mathbf{x}}|H_{t_n})\times \\ \nonumber
    &\exp\left( -\int_{\boldsymbol{V}}\int_{t_n}^t \lambda(u,{\mathbf{v}}|H_{t_n})du d \mathbf{v} \right)  d{\mathbf{x}}dt,  \\ \nonumber
    \mathbb{E}[{\mathbf{x}}_{n+1}|H_{t_n}] = &\int_{\boldsymbol{V}} {\mathbf{x}}\int_{t_n}^\infty \lambda(t,{\mathbf{x}}|H_{t_n})\times \\ \nonumber
    &\exp\left( -\int_{\boldsymbol{V}}\int_{t_n}^t \lambda(u,{\mathbf{v}}|H_{t_n})du d \mathbf{v} \right) dtd{\mathbf{x}},
\end{align}
where solving the integral in a continuous high-dimensional space could be computationally expensive, or inaccurate in the case of using sampling methods such as Monte Carlo sampling \citep{yeo2018rnn, hastings1970monte}. \textbf{\textit{Fourth}}, although (\ref{eq: firstmoment}) makes it feasible to predict the next event, the multi-event prediction demands sequential usage of the first-order moment, where at each step the history is updated via the most recent predicted event. This makes the problem of multi-event prediction challenging and highly prone to error accumulation.

To alleviate some of the mentioned shortcomings, previous works have proposed learning the intensity function $\lambda(t,{\mathbf{x}}|H_t)$ directly from data using neural networks. 
The early studies in this regard \citep{du2016recurrent,mei2017neural, xiao2017modeling} attempted to learn $\lambda(t,{\mathbf{x}}|H_t)$ using a recurrent neural network (RNN) and variants thereof. 
In a recent work, \cite{chen2020neural} designed a neural ordinary differential equation (ODE) architecture that can learn a continuous intensity function.
This is attained by combining jump and attentive continuous normalizing flows (CNFs)~\citep{chen2018neural} for space, and an RNN architecture for time. 
Despite being data-driven, these works do not tackle the last three drawbacks mentioned above. 
Moreover, due to the memory limitations of RNNs, these works are incapable of unveiling long-term dependencies. 
On the other hand, RNNs and their variants suffer from vanishing and exploding gradients, especially when applied to long sequential data. 
Furthermore, sequential training of both RNNs and jumpCNFs as proposed by \cite{chen2018neural} results in a slow and computationally expensive training process. 
Some other recent works \citep{zuo2020transformer,zhou2022neural}, proposed using attention mechanisms (already presented in the attentiveCNF in~\cite{chen2020neural}), to alleviate the second drawback mentioned earlier. 
The capability of discovering long-term dependencies while achieving fast training (via parallelization) is another benefit of this line of work. 
However, the last two drawbacks are still present in those approaches, hindering their utility for multi-event forecasting. 
Directly learning the temporal distribution using the variational autoencoders (VAEs)~\citep{kingma2014auto} augmented with CNFs~\citep{chen2018neural} as proposed by \cite{mehrasa2019point}, or utilizing the deep sigmoidal flow (DSF)~\citep{huang2018neural} and the sum of squares (SOS) polynomial flow~\citep{jaini2019sum} as proposed by \cite{shchur2019intensity} could tackle the first and third limitations. However, the second and last mentioned shortcomings remained to be solved.
In another recent work, \cite{zhu2021imitation} proposed learning the Hawkes intensity function based on Gaussian diffusion kernels while using imitation learning as a flexible model-fitting approach. However, their method could not tackle any of the aforementioned shortcomings.
Therefore, in the present work, we introduce a novel neural network that is capable of \emph{simultaneous} spatio-temporal multi-event forecasting while addressing all the above-mentioned shortcomings. 
To the best of our knowledge, this is the first work proposing a data-driven multi-event forecasting network for stochastic point processes. 
Our architecture augments the encoder and decoder blocks of a transformer architecture \citep{vaswani2017} from natural language processing (NLP) with probabilistic and bijective layers. 
This network design provides a batch of rich multi-event spatio-temporal distributions associated with multiple future events. 
We propose to directly learn/predict these distributions (without the need to learn the intensity function as opposed to the previous works), using self-supervised learning \citep{liu2021self}, marking a further point of departure with respect to existing work. 
The rest of this paper is organized as follows. 
In Section~\ref{sec:background}, we provide background knowledge on attention mechanisms and normalizing flows, since these are used as constituent blocks in our solution. 
Next, in Section~\ref{sec:networkall} we provide a formal definition of the problem to be solved along with our proposed neural multi-event forecasting network. Further, in Section~\ref{sec:experiments} we evaluate the performance of our network on a variety of datasets and compare it with baseline models. 
Finally, in Section~\ref{sec:conclusion} we finish the paper with conclusions and future goals. 
Our contributions can be summarized as follows:
\begin{itemize}
    \item We introduce a neural architecture that is capable of simultaneous multi-event forecasting of the time and location of discrete events in continuous time and space.
    \item We compare the performance of our solution with state-of-the-art models through extensive experiments on a variety of datasets that represent stochastic point events.
\end{itemize}
\vspace{-0.4cm}
\section{Background}
\label{sec:background}
\subsection{Attention and Transformers}
\label{sec: att}
Attention blocks play a critical role in the transformer architecture \citep{vaswani2017}. 
Here we introduce their fundamental pieces, already specializing them to the problem at hand.
Let $\{\boldsymbol{\kappa}_i = [t_i, {\mathbf{x}}_i,M_i]\}_{i=1:n}$ be a sequence of (column) vectors $\boldsymbol{\kappa}_i \in \mathbb{R}^{d+2}$ representing $n$ discrete events formed by concatenating the associated time $t_i \in \mathbb{R}$ and markers ${\mathbf{x}}_i \in \mathbb{R}^d$ and $M_i\in \mathbb{R}$.
The marker ${\mathbf{x}}_i$ denotes the location of the event at time $t_i$ in a $d$ dimensional space, and $M_i$ can be used to encode any other information of interest about the event, such as the magnitude of an earthquake or the biker's age. Even though we represent here $M_i$ as a scalar, these markers can also be higher dimensional.
Moreover, the events $\boldsymbol{\kappa}_i$ are temporally sorted such that $t_i < t_j$ for $i < j$.
Using learnable linear transformations, we form \textit{query}, \textit{key} and \textit{value} vectors (denoted by $\mathbf{q}_i$, $\mathbf{k}_i$, and $\mathbf{v}_i$, respectively) for each event $\boldsymbol{\kappa}_i$ as
\begin{align}
\label{eq:embedding}
    &\mathbf{q}_i = \mathbf{W}_q \, \boldsymbol{\kappa}_i, \enspace \enspace \enspace \enspace\enspace \mathbf{k}_i = \mathbf{W}_k\,\boldsymbol{\kappa}_i, \enspace \enspace \enspace \enspace \mathbf{v}_i = \mathbf{W}_v\,\boldsymbol{\kappa}_i,
\end{align}
where $\mathbf{W}_q \in \mathbb{R}^{d_k \times (d+2)}$, $\mathbf{W}_k \in \mathbb{R}^{d_k \times (d+2) }$, $\mathbf{W}_v \in \mathbb{R}^{d_v \times (d+2) }$.
For a given event $\boldsymbol{\kappa}_i$, we can build the matrix $\mathbf{K}_{(i)} = [\mathbf{k}_1 \mathbf{k}_2 \ldots \mathbf{k}_i]$, which contains as columns the key vectors of all the events up to (and containing) event $\boldsymbol{\kappa}_i$ and, similarly, the matrix $\mathbf{V}_{(i)} = [\mathbf{v}_1 \mathbf{v}_2 \ldots \mathbf{v}_i]$ values.
Based on this notation, we compute the hidden representation $\mathbf{h}_{i}$ of event $\boldsymbol{\kappa}_i$ as
\begin{align} 
\label{eq:att}
    \mathbf{h}_{i} = \mathbf{V}_{(i)} \,\, \mathrm{softmax} \left(  \frac{\mathbf{K}_{(i)}^\top \mathbf{q}_{i}}{\sqrt{d_k}} \right).
\end{align}
Note that in~(\ref{eq:att}) we are generating $\mathbf{h}_{i}$ by linearly combining the columns of $\mathbf{V}_{(i)}$, i.e., the values of all events preceding (and including) $\boldsymbol{\kappa}_i$.
The weights in this linear combination are given by a softmax applied to the inner products between the keys of every event and the query specific of $\boldsymbol{\kappa}_i$.
In this sense, $\mathbf{h}_{i}$ contains a weighted combination of the values of all preceding events, where these values and the weights in the combination can be learned by tuning the transformations $\mathbf{W}_q$, $\mathbf{W}_k$, and $\mathbf{W}_v$.

\subsection{Normalizing Flows}
\label{sec:NF}

A normalizing flow is a transformation of a simple probability distribution to a complex one using a sequence of differentiable and invertible bijective maps \citep{kobyzev2020normalizing}. 
Assume that $\boldsymbol{\kappa} \in \mathbb{R}^{d_\kappa}$ has a complex unknown distribution $p_{{\kappa}}$. 
Having $\mathbf{z} \in \mathbb{R}^{d_\kappa}$ as a random sample drawn from a simple known distribution $p_{{z}}$ (e.g., a normal distribution), we could use a sequence of invertible and differentiable functions $F = F_1 \circ F_2 \cdots \circ F_k$ to map $\mathbf{z}$ into  $\boldsymbol{\kappa}$ according to $\boldsymbol{\kappa} = F(\mathbf{z})$. 
Therefore, using the change of variables formula we have that
\begin{align}
\label{eq:NF}
     p_{{z}}(\mathbf{z}) = p_{{\kappa}}(F(\mathbf{z})) |\text{det}(\text{D}F(\mathbf{z})|,
\end{align}
where $\text{det}(\text{D}F(\mathbf{z}))$ denotes the determinant of the Jacobian matrix of $F(\mathbf{z})$ and accounts for the volume change in the density as we move from the simple distribution to the complex one. 
Equation~(\ref{eq:NF}) makes it feasible to have an explicit form for the complex underlying distribution $p_{{\kappa}}$ and we can effectively sample from it by generating samples from $p_z$ and transforming those via $F$.

\vspace{-0.3cm}
\section{Neural Multi-event Forecasting}
\label{sec:networkall}

Given a history of $n$ events, we want to determine the probability distribution (in time and space) of the next $L$ events.
Formally, we want to solve the following problem.

\noindent {\bf Problem 1.} Given the event history $H_{t_{n+1}}= \{(t_i, {\mathbf{x}}_i,M_i) \,| \,i=1:n\}$, estimate the probability distributions $\{p_l(t_l,{\mathbf{x}}_l|H_{t_{l}})\}_{l=n+1}^{n+L}$ of the next $L$ events.

We refer to the above problem as \emph{batched} or \emph{multi-event} prediction because given the history up to event $n$ we want to estimate the spatio-temporal probability densities of a \emph{batch} of $L$ events shown by $p_l$ in Problem 1. Note that despite using the word batched, these distributions do not need to be independent of each other.
We present here a data-driven solution of our problem where, during a training phase, we get to observe several sequences of length $n+L$ of our events of interest.
We propose a transformer-based architecture trained in a self-supervised manner to extract the relations between historical events.
Furthermore, we enhance the transformer block with probabilistic and bijective layers to construct history-dependent spatio-temporal distributions from the hidden states of the transformer.
Since $p_l(t_l, {\mathbf{x}}_l|H_{t_{l}}) = p_l({\mathbf{x}}_l | t_l, H_{t_{l}}) p_l(t_l|H_{t_{l}})$, we separately predict two batched history-dependent temporal \textit{and} spatial distributions. Figure~\ref{fig:network} shows a high-level view of our proposed network. 
Out of the $n+L$ events in the training sequences, $n$ are used as the history information and the inputs to the encoder, whereas the remaining $L$ are the inputs to the decoder in the training phase, and our goal (consistent with Problem~1) is to \emph{predict} their spatio-temporal distributions in the test phase. 

\vspace{-0.2cm}
\subsection{Network Architecture}
\label{sec:network}

\begin{figure*}[th]
    \centering
    \includegraphics[width=0.8\textwidth]{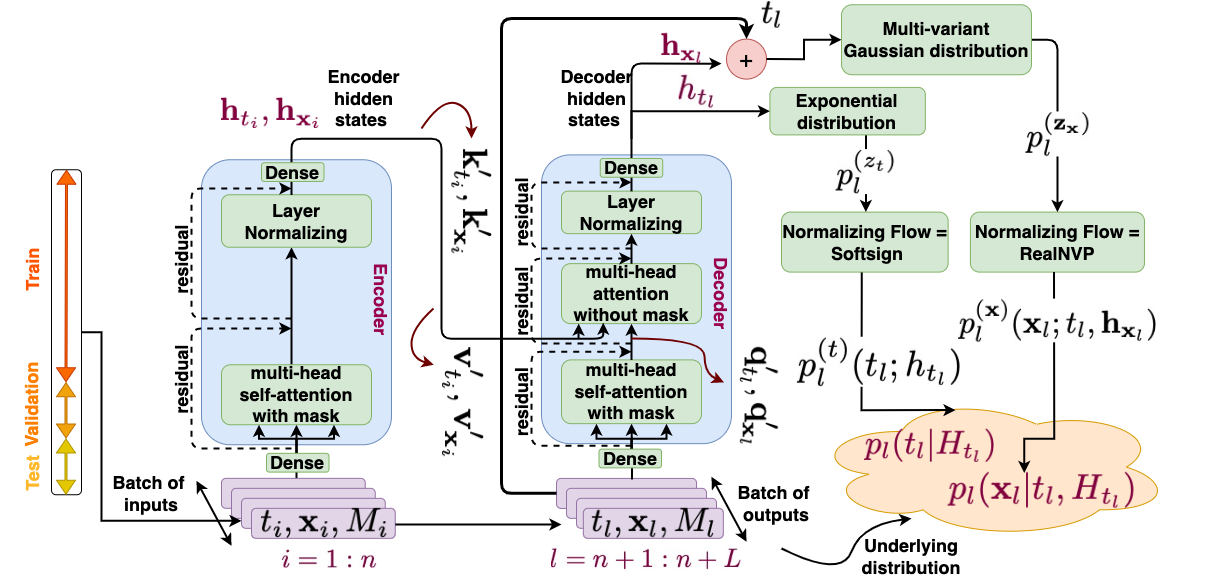}
    \caption{The framework of our proposed architecture. 
    Our network is composed of an encoder, a decoder, and two parallel batched probabilistic layers each followed by bijective layers known normalizing flows. 
    The input to the decoder is used to learn the output of the bijective layers using log-likelihood maximization in the training phase. 
    In the test phase, a sequence of all zeros is input to the decoder and the output of the bijective layers are the predicted batched probability distributions, i.e., our proposed solution to Problem~1.}
    \label{fig:network}
\end{figure*}

As shown in Figure~\ref{fig:network}, the transformer consists of two main blocks known as the encoder and the decoder. 
The encoder is fed with input sequence batches, where each sequence has $n$ events with their associated markers $(t_i, {\mathbf{x}}_i,M_i)$. 
Using multi-head attention blocks \citep{vaswani2017}, we extract representations for all $n$ events that capture history dependency. 
More precisely, via (\ref{eq:att}), we extract the hidden representation $\mathbf{h}_{t_i}$ for time and $\mathbf{h}_{{\mathbf{x}}_i}$ for space, separately, which will be used as part of the inputs to the decoder block.

The decoder contains two attention blocks that extract hidden representations of the output data (events $n+1$ through $n+L$) during the training phase. 
We denote these representations\footnote{Notice that, as a convention, we tend to use subindex $i$ (as in $\mathbf{h}_{t_i}$) to refer to the events in the interval $\{1, \ldots, n\}$ and subindex $l$ (as in $h_{t_l}$) for events in the interval $\{n+1, \ldots, n+L\}$. Also we use bold notation to show vector form of a varibale, meaning $h_{t_l}$ depicts 1 dimensional hidden states, whereas $\mathbf{h}_{t_i}$ stands for dimensions of higher than 1.} by $\{h_{t_l}, \mathbf{h}_{{\mathbf{x}}_l}\}_{l=n+1}^{n+L}$.  
The first attention block in the decoder captures history dependency within the subsequence of events between time $n+1$ and $n+L$.
We denote the outputs of this first block by $\mathbf{q}'_{t_l}$ and  $\mathbf{q}'_{{\mathbf{x}}_l}$ since these are used as queries for the second multi-head attention block in the decoder; see Figure~\ref{fig:network}.
The hidden states $\{\mathbf{h}_{t_i},\mathbf{h}_{{\mathbf{x}}_i}\}_{i=1}^n$ from the encoder are used as both the key and the value inputs to the second attention block in the decoder layer. 
In this way, the history dependency between the input (from event $1$ to $n$) and output (from event $n+1$ to $n+L$) events is encoded in the output hidden representations of the decoder, i.e., in $\{h_{t_l}, \mathbf{h}_{{\mathbf{x}}_l}\}_{l=n+1}^{n+L}$. 
Despite the typical implementation of the transformer decoder using look-ahead masks, due to our multi-event objective we remove this mask in the second attention block of the decoder. 
It is worth mentioning that, during testing, unlike the training phase, the time and markers for the events $n+1$ through $n+L$ are input as zero, so that no information is used for prediction beyond $H_{t_{n+1}}$, thus abiding by the constraint of Problem~1.
In Appendix~\ref{app: comparison} we explain the necessity behind using the time and markers information for the events $n+1$ through $n+L$ during the training phase, as opposed to the typical training strategy used in a conventional self-supervised transformer model. Moreover, in Appendix~\ref{app:ablation} we show how this information helps with the prediction results in a variety of different datasets.
Finally, normalizing, residual, and dense layers are used in both the encoder and the decoder blocks for dimension matching and enhancing the training. 

In our next step, we separately inject $\{h_{t_l}\}_{l=n+1}^{n+L}$ and $\{\mathbf{h}_{{\mathbf{x}}_l}\}_{l=n+1}^{n+L}$ into two separate batched probabilistic layers~\citep{durr2020probabilistic}. 
In essence, for the time variable $t_l$, we define a trainable map going from $h_{t_l}$ into the parameter of an exponential distribution.
We denote this distribution by $p_l^{(z_t)}( \, \cdot \, ; h_{t_l})$.
Similarly, for the spatial variable $\mathbf{x}_l$, we define a trainable map going from $(\mathbf{h}_{{\mathbf{x}}_l}, t_l)$ into the parameters of a multivariate Gaussian distribution.
We denote this distribution by $p_l^{(\mathbf{z}_{{\mathbf{x}}})} ( \, \cdot \,; t_l, \mathbf{h}_{{\mathbf{x}}_l})$.
We learned the parameters of these distributions independently for every $l \in \{ n+1, \ldots, n+L\}$.
However, it should be noticed that the hidden states $\{h_{t_l}\}_{l=n+1}^{n+L}$ and $\{\mathbf{h}_{{\mathbf{x}}_l}\}_{l=n+1}^{n+L}$ already encode spatio-temporal dependencies, so that the parameters learned for the probability distributions of different events already capture the correlations with other events across time.
Also notice that the parameters of the spatial distribution depend on the time instant $t_l$ since we want to estimate the joint probability density over time and space (see Problem 1) as $p_l(t_l, {\mathbf{x}}_l|H_{t_{l}}) = p_l({\mathbf{x}}_l | t_l, H_{t_{l}}) p_l(t_l|H_{t_{l}})$.\footnote{Note that during the test phase we only feed the decoder with $t_l = 0$ when modeling $p_l^{(z_t)}( \, \cdot \, ; h_{t_l})$; However, we use the true known time values $t_{l_{\text{known}}}$ as inputs to the probablistic layers when modeling $p_l^{(\mathbf{z}_{{\mathbf{x}}})} ( \, \cdot \,; t_l, \mathbf{h}_{{\mathbf{x}}_l})$ in both train and test phases. More details are provided in Appendix~\ref{app:datanorm}.}

Although the probabilistic layers have a critical role in forming the underlying batched distributions, they can only take simple and tractable forms (in our case, exponential and multivariate Gaussian), which are not expressive enough to capture the true underlying distributions of the times and locations of future events. 
Thus, a set of bijective layers, known as normalizing flows \citep{kobyzev2020normalizing}, are used to transform them into richer and more complex batched forms. 
As discussed in Section~\ref{sec:NF}, we can use (\ref{eq:NF}) to address the transformation of distributions $p_l^{(z_t)}$ and $p_l^{(\mathbf{z}_{{\mathbf{x}}})}$ into more complex forms. 
We use the exponential and multivariate normal distributions as the base distributions $p_l^{(z_t)}$ and $p_l^{(\mathbf{z}_{{\mathbf{x}}})}$ because they have been used extensively as parametric kernels to form the Hawkes intensity function in (\ref{eq:lambda}) \citep{ogata1998space, ogata1988statistical}.
Moreover, in Appendix~\ref{app:hyperparams} we have shown that the histograms depicting the space and time distributions in various datasets are not too far from being a multivariate Gaussian distribution for space and an exponential for consecutive time intervals. In this sense, normalizing flows only need to learn to further adjust these distributions to the observed data.
More precisely, we find continuous and invertible functions $F_1$ and $F_2$ such that, for every $l$, these functions can transform samples from $p_l^{(z_t)}$ and $p_l^{(\mathbf{z}_{{\mathbf{x}}})}$ (denoted by $z_{t_l}$ and $\mathbf{z}_{{\mathbf{x}_l}}$) into $t_l$ and $\mathbf{x}_l$ as in
\begin{align}
    t_l = F_1(z_{t_l}),  \enspace \enspace  {\mathbf{x}}_l = F_2(\mathbf{z}_{{\mathbf{x}_l}}). 
\end{align}
Using (\ref{eq:NF}) we can relate the known expressions of the base distributions with the distributions of $t_l$ and ${\mathbf{x}}_l$ using the Jacobians of $F_1$ and $F_2$, respectively.
We denote these learned distributions by $p_l^{(t)}( \, \cdot \, ;h_{t_l})$ and $p_l^{({\mathbf{x}})}( \, \cdot \, ;t_l, \mathbf{h}_{{\mathbf{x}}_l})$.
In terms of parametric functional forms, we use a softsign bijector for $F_1$ and a RealNVP bijector \citep{dinh2016density} for $F_2$.
To train these bijections, we seek to maximize the log-likelihood of the true events $\{t_l, {\mathbf{x}}_l\}_{l=n+1}^{n+L}$ under the learned distributions $p_l^{(t)}( \, \cdot \, ;h_{t_l})$ and $p_l^{({\mathbf{x}})}( \, \cdot \, ;t_l, \mathbf{h}_{{\mathbf{x}}_l})$.
More details about the training of our network will be discussed in Section~\ref{sec: training} and Appendix~\ref{app:networkdetails}.
Going back to Problem~1, by maximizing the log-likelihood we are training our network in an end-to-end fashion such that $p_l^{(t)}( t_l ;h_{t_l}) \approx p_l(t_l|H_{t_l})$ and $p_l^{({\mathbf{x}})}( {\mathbf{x}}_l ;t_l, \mathbf{h}_{{\mathbf{x}}_l}) \approx p_l({\mathbf{x}}_l|t_l, H_{t_l})$, where, we recall, $H_{t_l}$ contains the history of events up to (but not including) event $l$.
From here we can obtain 
\begin{equation}\label{eq:sol_problem}
p_l(t_l, {\mathbf{x}}_l|H_{t_l}) = p_l({\mathbf{x}}_l|t_l, H_{t_l}) p_l(t_l|H_{t_l}), 
\end{equation}
where $l = n+1, \ldots, n+L$.
Finally, be recalling that during testing every event beyond $n$ is completed via zero padding. Thus in~(\ref{eq:sol_problem}) we are effectively obtaining a distribution for time and space of every event $l$ in the range $n+1$ to $n+L$ given $H_{t_{n+1}}$, as we aimed for in Problem~1.

\begin{table*}[th]
\caption{Results representing the negative log-likelihood for the output events in the range $n+1$ to $n+L$ for the learned distributions (less is better). $\pm$ indicates the standard deviation of the loss among all test batch sequences.}
\label{table:result}
\vskip 0.15in
\begin{center}
\begin{small}
\begin{sc}
\begin{tabular}{lcccr}
\toprule
& Earthquake & Citibike &Covid-19& Pinwheel \\
\midrule
Homo-Poisson $(p_l^{(t)})$       & $6.91_{\pm 0.0}$ & $6.83_{\pm 0.0}$ & $6.32_{\pm 0.0}$ & $6.82_{\pm 0.0}$   \\
Hawkes $(p_l^{(t)})$          & $0.42_{\pm 0.09}$ & $-0.53_{\pm 0.05}$ & $-3.79_{\pm 0.01}$ & $\mathbf{-3.98_{\pm 1.54}}$\\
Self-correcting $(p_l^{(t)})$         & NAN & $1.95_{\pm 0.11}$ & $4.6_{\pm 0.02}$ & $2.03_{\pm 0.73}$
\\ \hline \\
Conditional-GMM $(p_l^{({\mathbf{x}})})$  &$4.88_{\pm 0.29}$ &$3.05_{\pm 1.26}$  &$3.48_{\pm 1.37}$  &$4.38_{\pm 0.42}$
\\ \hline \\
\bf Our network ($p_l^{(t)}$)   & $\mathbf{0.33_{\pm 0.17}}$ & $\mathbf{-6.279_{\pm  0.123}}$ & $\mathbf{-5.87_{\pm  0.09}}$ & $1.27_{\pm 0.26}$\\
\bf Our network ($p_l^{({\mathbf{x}})}$)   & $\mathbf{2.03_{\pm 0.26}}$ & $\mathbf{-3.28_{\pm 0.14}}$ & $\mathbf{2.18_{\pm 0.21}}$ & $\mathbf{1.57_{\pm 0.09}}$\\
\bf Our network ($p_l^{(t, {\mathbf{x}})}$)   & $\mathbf{2.36_{\pm 0.36}}$ & $\mathbf{-9.56_{\pm 0.18}}$ & $\mathbf{-3.7_{\pm 0.16}}$ & $\mathbf{2.84_{\pm 0.19}}$ \\
\bottomrule
\end{tabular}
\end{sc}
\end{small}
\end{center}
\vskip -0.1in
\end{table*}

\vspace{-0.3cm}
\subsection{Addressing the Limitations of Existing Approaches}
\label{sec:limitations}

As previewed in Section~\ref{sec: intro}, our architecture in Figure~\ref{fig:network} was inspired by the need of addressing four shortcomings present in classical Hawkes processes and partially shared by more modern alternatives.
Here, we explicitly discuss how the different constituent blocks in our architecture address these deficiencies.
\textbf{\textit{First}}, we do not depend on prespecified kernels to model the probability densities of time and location.
Notice that, although we adopt exponential and Gaussian distributions for the latent variables $z_{t_l}$ and $\mathbf{z}_{{\mathbf{x}_l}}$, these are later transformed via normalizing flows to better represent the observed data.We provide a mathematical description on this in Appendix~\ref{app:network}.
\textbf{\textit{Second}}, the attention mechanisms in both the encoder and decoder blocks of the transformer can learn heterogeneous effects between events across time.
\textbf{\textit{Third}}, the explicit forms of $p_l^{(t)}( t_l ;h_{t_l})$ and $p_l^{({\mathbf{x}})}( {\mathbf{x}}_l ;t_l, \mathbf{h}_{{\mathbf{x}}_l})$ obtained via normalizing flows facilitate the fast sampling of future events via (\ref{eq:sol_problem}) without having to rely on integration as in~(\ref{eq: firstmoment}).
\textbf{\textit{Fourth}}, the incorporation of the decoder block of the transformer enables multi-event prediction bypassing the need of sequential one-by-one event prediction.
To be more precise, the use of our decoder block implies that, \emph{during training}, we are incorporating the future information $\{h_{t_l}, \mathbf{h}_{{\mathbf{x}}_l}\}_{l=n+1}^{n+L}$ along with their dependence on the encoded history $\{\mathbf{h}_{t_i}, \mathbf{h}_{{\mathbf{x}}_i}\}_{i=1}^{n}$. 
This makes our training procedure self-supervised rather than unsupervised, unlike all previous works in the area. 
In essence, during training, the network learns to fit the distribution of multiple future events jointly.
Hence, during testing, we can zero-pad the input to the decoder and still obtain an estimation of the spatio-temporal probability densities of several events in the future. A detailed description of our network training as a self-supervised model is given in Appendix~\ref{app: comparison}.

\vspace{-0.4cm}
\section{Experiments}
\label{sec:experiments}
\vspace{-.1cm}
\subsection{Data}
\label{sec: data}
\vspace{-.1cm}
We used a variety of synthetic and real-world datasets representing point processes with discrete events. 
We briefly introduce these datasets here; more details can be found in Appendix~\ref{app:preprocessing}. 
In all datasets, we collected sequences of 500 events with an overlap of 498 events.
We used $n=497$ events in each sequence as the inputs and the final $L=3$ events as the outputs that we want to predict during the testing phase. Train, validation, and test sets are formed according to the $80\%-14\%-6\%$ split rule after shuffling the formed sequences.

\textbf{South California Earthquakes.} 
Earthquake events from 2008 to 2016 in South California \citep{ross2019searching} of magnitude at least 2.5.
The event description includes time, location in $3$ dimensions, magnitude and consecutive events' time intervals as event markers. 

\textbf{Citibike.} Rental events from a bike sharing service in New York City \citep{Amazoncitibike}. The event description includes starting time of the biking, location in $2$ dimensions, the biker's birth year, and consecutive rentals' time intervals as event markers.

\textbf{Covid-19.} Daily Covid-19 cases in different states of the United States~\citep{TheNewYorkTimes}. 
The event description includes day of catching Covid-19, location in $2$ dimensions, the number of cases on that day, and consecutive events' time intervals as event markers.

\textbf{Pinwheel.} Hawkes pinwheel dataset introduced in~\cite{chen2020neural}. 
Hawkes time instances were simulated using the thinning algorithm~\citep{ogata1981lewis} and assigned to a cluster-based pinwheel distribution. 
For simplicity, we assigned the same magnitude to all spatial points sampled from the formed distribution when forming the data sequences. 

\vspace{-0.3cm}
\subsection{Baselines}
\label{sec:baselines}

We consider two categories of baseline methods, namely, those that try to predict the time and those that try to predict the location of the next event.
Time category models are used to learn the intensity function from input times and we predict the expected time of multiple events in the future by sequentially utilizing (\ref{eq: firstmoment}).
These models include the Hawkes process (also known as the ETAS model~\citep{ogata1988statistical,ogata1998space}), self-correcting point process~\citep{isham1979self}, and the homogeneous Poisson process~\citep{pasupathy2010generating}. 
Note that the sequential prediction of the future time-points for more than one event, in cases of the Hawkes and self-correcting intensities, relies on a redefined history that incorporates the very last predicted event time in the previous step.
We also use space models to sequentially learn the space distributions of multiple events in the future. 
In this regard, we use the conditional multi-variate Gaussian mixture model (GMM)~\citep{murphy2012machine, bishop2006pattern} to learn the space distribution, where the Gaussian kernel parameters are learned from the historical events. 
All other works mentioned in Section~\ref{sec: intro} aimed to learn the Hawkes process intensity function without approaching multi-event forecasting.
Hence, we restrict our comparison to the aforementioned classical baselines.

\begin{figure*}[th]
    \centering
    \includegraphics[width=0.85\textwidth]{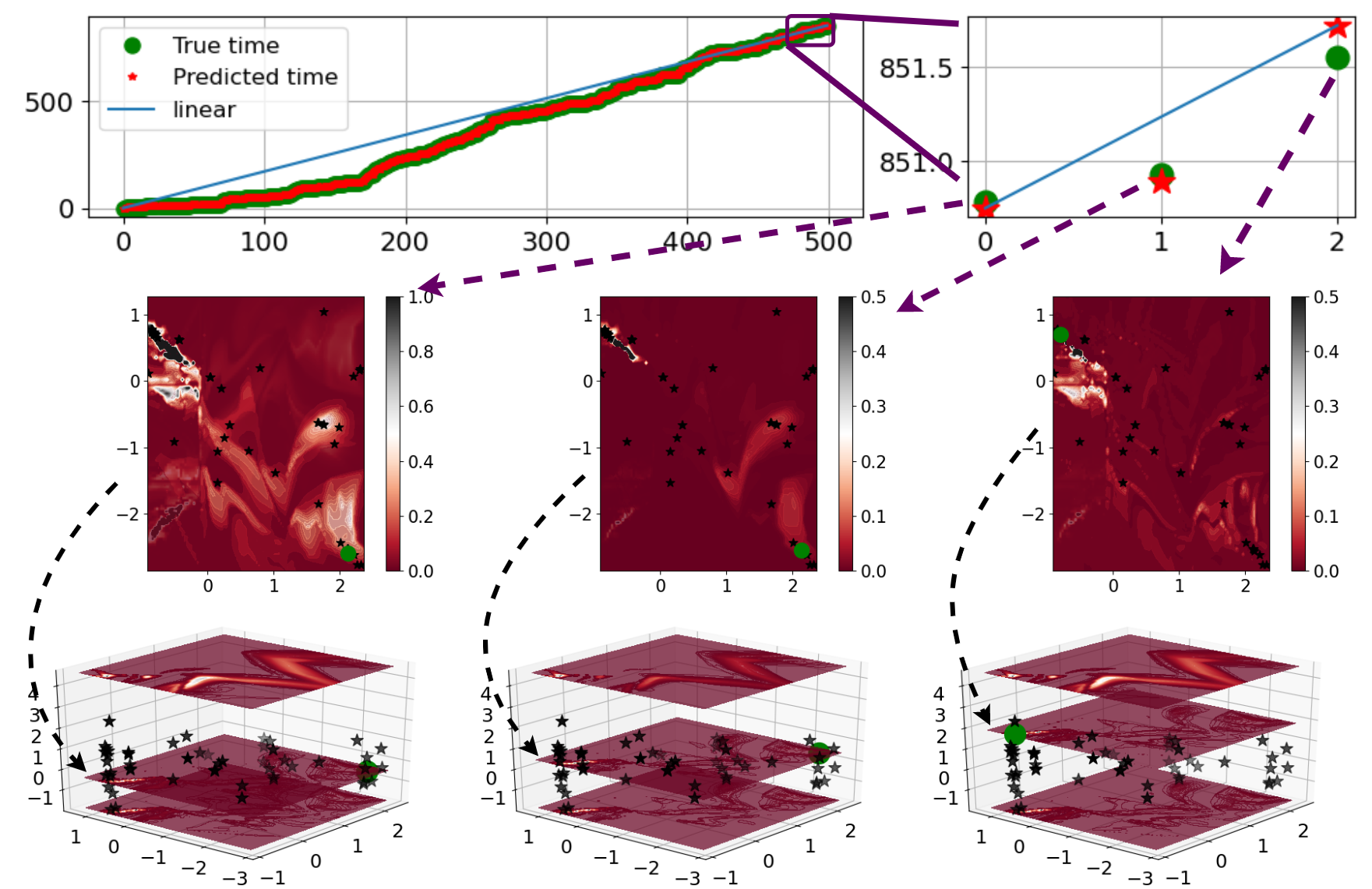}
    \caption{Visualizations of the learned history-dependent batched spatial and temporal distributions for the last three events of an unseen earthquake sequence using \textbf{our proposed architecture}. The red stars on the \textbf{top} figures are the simultaneous prediction of the true events' time, shown by green circles, where we took an average between 1000 samples taken from the output predicted time distributions.
    The \textbf{bottom} figures depict the predicted batched spatial distribution associated with the last three events in 3 dimensions where the z-axis is along events' depth. 
    The 2-d representation of the last three events' predicted spatial distributions mapped to the future events' depth are shown on the \textbf{middle} plots. In these figures we have the predicted density on the left (i.e., for the first predicted event) and the other two figures depict the consecutive predicted density differences for the two subsequent events.
    The black stars on the spatial distributions represent the last 50 events of the input sequence as the history information.}
    \label{fig:earthtransformer}
\end{figure*}

\vspace{-0.2cm}
\subsection{Network Train and Test}
\label{sec: training}

Training starts by passing batches of input and output sequences to the encoder and the decoder, respectively. 
The encoder outputs are also passed to the decoder as earlier explained in Section~\ref{sec:network}. 
We used inputs and outputs with all the existing markers for each dataset as mentioned in Section~\ref{sec: data}. 
A list of the hyperparameters used in this work is shown in Table~\ref{table: hyper-params}.
We define the loss as minimizing the negative log-likelihood 
\begin{align}
\label{eq:losstransformer}
    \text{loss} &= -\sum_{l=n+1}^{n+L}\log p_l^{(t)}(t_l;h_{t_l}) -\sum_{l=n+1}^{n+L}\log p_l^{({\mathbf{x}})}({\mathbf{x}}_l;t_l, \mathbf{h}_{{\mathbf{x}}_l}).
\end{align}
Thus, we are effectively learning the coefficients in the transformer blocks -- which affect the hidden representations $h_{t_l}$ and $\mathbf{h}_{{\mathbf{x}}_l}$ -- and the coefficients in the probabilistic and bijective layers -- which parameterize $p_l^{(t)}$ and $p_l^{({\mathbf{x}})}$ -- to maximize the likelihood of generating the observed data.

The baseline methods are trained to encode the history of past events and to sequentially predict the expected next event based on (\ref{eq: firstmoment}). 
Therefore, in each step we compute the expected time $\hat{t}_l$ and location $\hat{{\mathbf{x}}}_l$ of event $l$ and we want these to be close to the true values $t_l$ and ${\mathbf{x}}_l$ before updating the history for the next prediction step. 
To satisfy this condition, during the training phase, we add regularizers to the negative log-likelihood as in 
\begin{align}
\label{eq:loss}
    \text{loss} = &-\sum_{l=1}^{n+L}\log p_{t}(t_l|H_{t_l}) -\sum_{l=1}^{n+L}\log p_{\mathbf{x}}({\mathbf{x}}_l|t_l, H_{t_l})\\ \nonumber
    &+ \lambda_1\sum_{l=n+1}^{n+L}|t_l-\hat{t}_l| + \lambda_2\sum_{l=n+1}^{n+L}||{\mathbf{x}}_l-\hat{{\mathbf{x}}}_l||_2 ,
\end{align}
where $p_{t}(\, \cdot \,|H_{t_l})$ and $p_{\mathbf{x}}(\, \cdot \,|t_l, H_{t_l})$ are the learned probability densities for the baseline method at hand, and $\lambda_1$ and $\lambda_2$ encode the relative weights of the regularizers. 
To be more precise, if a method only predicts the time, we use only the first and third terms in~(\ref{eq:loss}) as the loss and if a method only predicts the location we use the second and fourth terms as the loss.
For a fair comparison between the loss of our network vs. the baselines, during the testing phase we report the negative log-likelihood for events $n+1$ through $n+L$ for all the tested models, where we also remove the regularizers from~(\ref{eq:loss}). 
We train our network using the Adam optimizer with an initial learning rate of $0.001$ and a scheduled learning rate rule. 
We used droupout layers to reduce overfitting and ran $1000$ epochs using batches of size $32$.\footnote{Code for preprocessing and training are open-sourced at \url{https://github.com/Negar-Erfanian/Neural-spatio-temporal-probabilistic-transformers}}

\begin{figure*}[th]
    \centering
    \includegraphics[width=0.84\textwidth]{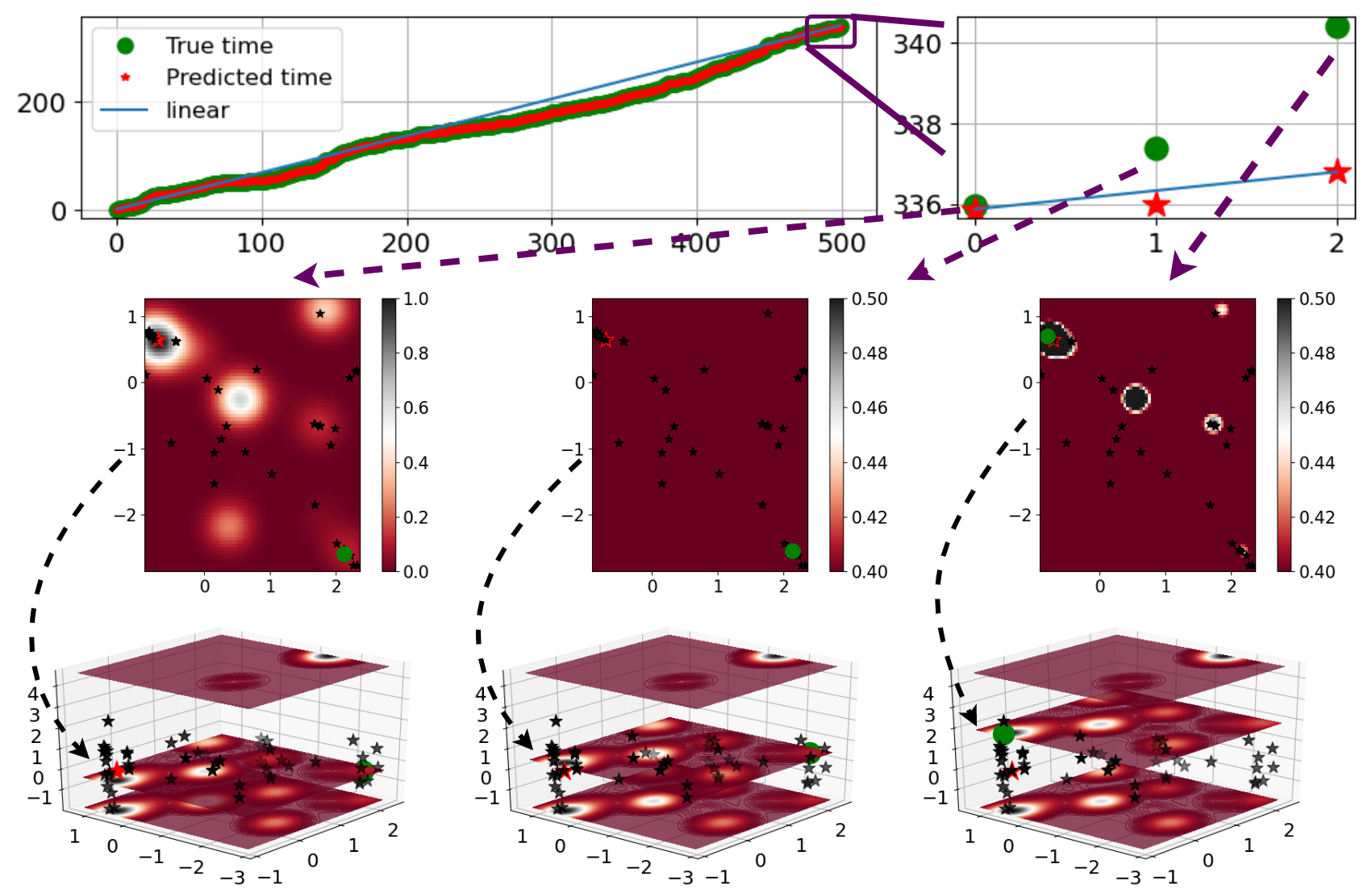}
    \caption{Same visualizations as Figure~\ref{fig:earthtransformer} for the same data using \textbf{Hawkes model} for time, augmented with \textbf{conditional-GMM} for space. Here the prediction in time and space is performed sequentially using (\ref{eq: firstmoment}) by updating the history at each prediction step.}
    \label{fig:earthbench}
\end{figure*}

\subsection{Results}
\label{sec:results}

\vspace{-.15cm}
Figure~\ref{fig:earthtransformer} shows the predicted batched spatial and temporal distributions for the next three events on the South California Earthquake data \citep{ross2019searching} using our proposed architecture. 
On the top right figure, we took the average of 1000 samples taken from the predicted batched history dependent time distributions to forecast the time of future events shown by the red stars. 
The linear trend is shown to emphasize that our network is capable of predicting time events with different in-between time-intervals. 
The bottom figures show the 3-dimensional representation of batched history-dependent multi-modal predicted densities along the main faults, prone to earthquake occurrence. 
In the 3-d visualization we show three slices along the z-axis, representing the lowest, highest, and the depth associated with the true future event, shown by the green circle. 
In the middle, we show the 2-dimensional density visualizations that are associated with the true event's depth shown in the 3-d figures below them.
Note that, while the bottom figures represent the predicted densities, in the middle figures we have the predicted density on the left (i.e., for the first predicted event) and the other two middle figures depict the consecutive predicted \emph{density differences} for the two subsequent events. 
This is done to highlight the performance of our network in predicting simultaneous but different multi-event densities. 
Each predicted space density is associated with the assigned events' predicted time shown above it. 
Despite seeing only slight differences between the predicted space densities, regions that are most expected to have the next event are still well-recognized. 
Note that while the same history is encoded for all the three future events, the differences between the predicted batched densities are highly influenced by the decoder layer which incorporates future information that was available during the training phase. 
Considering that consecutive outputs in most sequences occur very close to each other in time and space, the decoder layer is trained to represent the same behaviour in the test phase, predicting batched densities with slight but informative differences. 
The high bias on the upper left of the space density maps is caused by the highly earthquake-prone regions/faults.

\vspace{-.1cm}
Figure~\ref{fig:earthbench} on the other hand, represents the sequential prediction performance of the baseline models on the same dataset used in Figure~\ref{fig:earthtransformer}. 
The Hawkes model and the conditional GMM model are used to learn the time and space distributions, respectively, where we used (\ref{eq: firstmoment}) to sequentially predict the time and location of the next event. 
Comparing this result with Figure~\ref{fig:earthtransformer} we see better performance of our network in predicting the time and space distribution of future events on earthquake data. 
Moreover, the conditional GMM model is incapable of predicting a complex and well-defined density along fault lines representing earthquake-prone regions. 
More details on how we perform sequential prediction is given in Appendix~\ref{app:baseline}. 
The black stars on spatial density plots in Figures~\ref{fig:earthtransformer} and \ref{fig:earthbench} show the last 50 out of the 497 historical events that are used as the encoder inputs.

The results shown in Table~\ref{table:result} for all the experiments indicate that, except for the Hawkes time pinwheel synthetic dataset, our network outperforms all the baseline models. 
Due to a very high error accumulation associated with the sequential prediction using the conditional GMM model, we used the true rather than the predicted values at each prediction step to update the history. 
More visualization results are given in Appendix~\ref{app:analysis}. 
Since the times associated with the pinwheel dataset are synthetically simulated from the Hawkes model, Hawkes is a better candidate for forecasting the expected time distribution for this dataset as shown in Table~\ref{table:result}. 
Among the given baseline models, the self-correcting and homogeneous Poisson are performing the worst, where the self-correcting with the inhibitory assumption behind history dependency never converged on the earthquake dataset. 
While all the provided baseline models assume certain parametric forms for the space or time distributions, our proposed network is applicable to any discrete type dataset even in the absence of underlying physics or known models that can guide the right choice of a spatio-temporal kernel. 

\vspace{-.2cm}
\section{Conclusion}
\label{sec:conclusion}
\vspace{-.1cm}
We have proposed a novel probabilistic approach by enhancing a transformer architecture to conditionally learn the spatio-temporal distribution of multiple stochastic discrete events. 
Our network combines the transformer model and normalizing flows augmented with batched probabilistic layers to simultaneously learn the underlying distributions of multiple future events using self-supervised learning. 
The attention blocks of the transformer architecture assign score-based history dependency among events, which can capture both excitatory and inhibitory behaviors while being parallel and fast to train.
We show that our approach achieves state-of-the-art performance on different spatio-temporal datasets collected from a wide range of domains.

\vspace{-.1cm}
For future work, we are interested in using other available sources of data as proposed by the work of \cite{okawa2022context} such as the GPS data and under-earth images to propose a context-aware batched spatio-temporal distribution forecasting tool.
Another interesting direction is to use the diffusion model~\citep{song2020denoising, ho2020denoising, nichol2021improved} instead of normalizing flows as generative models to learn the probability distributions. 
We hope that by adding gradual noise in the reverse diffusion denoising steps, we can enhance learning regions associated with rare events.


\section*{Acknowledgements}
Maarten de Hoop was supported by the Simons Foundation under the MATH + X program, the National Science Foundation under grant DMS-2108175, and the corporate members of the Geo-Mathematical Imaging Group at Rice University.

\bibliography{example_paper}
\bibliographystyle{icml2023}

\newpage
\appendix
\onecolumn
\section{Further Details on Our Proposed Network Architecture and Training}
\label{app:networkdetails}
\subsection{Decoder Layer}
\label{app: dec}
In this section, we further expand the description on how the hidden representations $\{h_{t_l}, \mathbf{h}_{{\mathbf{x}}_l}\}_{l=n+1}^{n+L}$ are formed from the decoder block. 
As earlier we mentioned in Section~\ref{sec: att}, $\{\mathbf{h}_{t_i}, \mathbf{h}_{{\mathbf{x}}_i}\}_{i=1}^{n}$ are formed using (\ref{eq:att}) and will be later used as the keys $\{\mathbf{k}'_{t_i}, \mathbf{k}'_{{\mathbf{x}}_i}\}_{i=1}^{n}$ and values $\{\mathbf{v}'_{t_i}, \mathbf{v}'_{{\mathbf{x}}_i}\}_{i=1}^{n}$ for the second attention layer in the decoder block. However in the training phase, the queries $\{\mathbf{q}'_{t_l}, \mathbf{q}'_{{\mathbf{x}}_l}\}_{l=n+1}^{n+L}$ to the mentioned attention block are formed as follows. 
Let $\{\boldsymbol{\kappa}_l = [t_l, {\mathbf{x}}_l,M_l]\}_{l=n+1}^{n+L}$ be a sequence of (column) vectors $\boldsymbol{\kappa}_l \in \mathbb{R}^{d+2}$ representing $L$ discrete events formed by concatenating the associated time $t_l \in \mathbb{R}$ and markers ${\mathbf{x}}_l \in \mathbb{R}^d$ and $M_l\in \mathbb{R}$.
Even though we represent here $M_l$ as a scalar, these markers can also be higher dimensional.
Moreover, the events $\boldsymbol{\kappa}_l$ are temporally sorted such that $t_l < t_k$ for $l < k$.
Using (\ref{eq:embedding}) we form query $\mathbf{q}_l \in \mathbb{R}^{d_k'} $, key $\mathbf{k}_l \in \mathbb{R}^{d_k'}$ and value $\mathbf{v}_l \in \mathbb{R}^{d_v'}$ vectors for each event $\boldsymbol{\kappa}_l$.
For a given event $\boldsymbol{\kappa}_l$, we build the matrix $\mathbf{K}_{(l)} = [\mathbf{k}_{n+1} \mathbf{k}_{n+2} \ldots \mathbf{k}_l]$, which contains as columns the key vectors of all the events up to (and containing) event $\boldsymbol{\kappa}_l$ and, similarly, the value matrix $\mathbf{V}_{(l)} = [\mathbf{v}_{n+1} \mathbf{v}_{n+2} \ldots \mathbf{v}_l]$.
Based on this notation, using (\ref{eq:att}) we compute the hidden representation $\mathbf{h}_{l}$ of event $\boldsymbol{\kappa}_l$. These hidden representations, which are the outputs from the first attention layer in the decoder block, are now used as the queries $\{\mathbf{q}'_{l}\}_{l=n+1}^{n+L}$ to the second attention layer in the decoder block as shown in Figure~\ref{fig:network}. Note that in our setup we form these queries for both time and space in parallel using separate weights. This will give us the freedom to tune the temporal and spatial information separately while using the same input information. 
Therefore we have 
$\{\mathbf{q}'_{t_l}\}_{l=n+1}^{n+L}$ as the temporal query and $\{ \mathbf{q}'_{{\mathbf{x}}_l}\}_{l=n+1}^{n+L}$ as the spatial query.
Having $\{\mathbf{q}'_{t_l}, \mathbf{q}'_{{\mathbf{x}}_l}\}_{l=n+1}^{n+L}$, $\{\mathbf{k}'_{t_i}, \mathbf{k}'_{{\mathbf{x}}_i}\}_{i=1}^{n}$ and $\{\mathbf{v}'_{t_i}, \mathbf{v}'_{{\mathbf{x}}_i}\}_{i=1}^{n}$ respectively as the queries, keys and values to the second attention layer in the decoder block, we repeat the procedure described above to compute $\{h_{t_l}, \mathbf{h}_{{\mathbf{x}}_l}\}_{l=n+1}^{n+L}$ as the temporal and spatial hidden representations that will be further injected into the following probabilistic layers. 
Note that while dimensions $d_k'$ and $d_v'$ can be different, for simplicity in our computations we set $d_k'= d_v'$. 
Also notice that we use feed forward layers to set $\{h_{t_l} \in \mathbb{R}, \mathbf{h}_{{\mathbf{x}}_l}\in \mathbb{R}^{d}\}_{l=n+1}^{n+L}$ where d denotes the space dimension associated with the dataset. This is due to the fact that the following bijective layers do not allow change of dimension while mapping. 
\subsection{Probabilistic and Bijective Layers}
\label{app:network}
In Section~\ref{sec:network}, we showed the usage of two individual batched probabilistic layers that are separately fed with $\{h_{t_l}\}_{l=n+1}^{n+L}$ and $\{(\mathbf{h}_{{\mathbf{x}}_l}, t_l)\}_{l=n+1}^{n+L}$, where $\{h_{t_l}, \mathbf{h}_{{\mathbf{x}}_l}\}_{l=n+1}^{n+L}$ are formed as described in Appendix~\ref{app: dec} and $\{t_l\}_{l=n+1}^{n+L}$ are the known time values. We denoted the outputs of these layers by $p_l^{(z_t)}( \, \cdot \, ; h_{t_l})$ and $p_l^{(\mathbf{z}_{{\mathbf{x}}})} ( \, \cdot \,; t_l, \mathbf{h}_{{\mathbf{x}}_l})$ as the conditional exponential and multivariate Gaussian distributions, respectively, whose parameters are learned from $\{h_{t_l}\}_{l=n+1}^{n+L}$ and $\{(\mathbf{h}_{{\mathbf{x}}_l}, t_l)\}_{l=n+1}^{n+L}$. To further expand the mathematical description of these layers' outputs we have

\begin{align}
\label{eq:problayers}
    &p_l^{(z_t)}( \, \cdot \, ; h_{t_l}) = \frac{1}{\beta_l}\exp(-\frac{z_{t_l}}{\beta_l}), \nonumber \\ &p_l^{(\mathbf{z}_{{\mathbf{x}}})} ( \, \cdot \,; t_l, \mathbf{h}_{{\mathbf{x}}_l}) = ((2\pi)^{n_{\mathbf{z}_{{\mathbf{x}}_l}}}|\Sigma_l|)^{-\frac{1}{2}}\exp(-\frac{1}{2}({\mathbf{z}_{{\mathbf{x}}_l}} - \mu_l)^{\top}\Sigma_l^{-1}({\mathbf{z}_{{\mathbf{x}}_l}} - \mu_l)),
\end{align}

where $\beta_l$ is learned as a function of $h_{t_l}$, and $\{\Sigma_l, \mu_l\}$ are learned as functions of $\{t_l, \mathbf{h}_{{\mathbf{x}}_l}\}$.
$p_l^{(z_t)}( \, \cdot \, ; h_{t_l})$ and $p_l^{(\mathbf{z}_{{\mathbf{x}}})} ( \, \cdot \,; t_l, \mathbf{h}_{{\mathbf{x}}_l})$ in (\ref{eq:problayers}) are individually followed by a softsign bijector $F_1$ and a RealNVP bijector $F_2$~\citep{dinh2016density}, respectively, as two separate bijective layers to model the desired batched conditional temporal and spatial distributions $p_l^{(t)}( \, \cdot \, ;h_{t_l})$ and $p_l^{({\mathbf{x}})}( \, \cdot \, ;t_l, \mathbf{h}_{{\mathbf{x}}_l})$. 
Using (\ref{eq:NF}), we have
\begin{align}
    &p_l^{(t)}( \, \cdot \, ;h_{t_l}) = p_l^{(z_t)}( \, \cdot \, ; h_{t_l})|\text{det}(\text{D}F_1(z_{t_l})|^{-1}, \nonumber\\
    &p_l^{({\mathbf{x}})}( \, \cdot \, ;t_l, \mathbf{h}_{{\mathbf{x}}_l}) = p_l^{(\mathbf{z}_{{\mathbf{x}}})}( \, \cdot \,; t_l, \mathbf{h}_{{\mathbf{x}}_l})|\text{det}(\text{D}F_2(\mathbf{z}_{{\mathbf{x}}_l})|^{-1}.
\end{align}
Notice that the usage of the softsign bijective function $F_1(y) = \frac{y}{|y|+1}$ after the exponential probabilistic layer restricts the outputs between $0$ and $1$, i.e. $F_1: (0, \infty) \rightarrow (0,1)$. 
This is consistent with the fact that we always convert the time values $\{t_i\}_{i = 1}^{n+1}$ and $\{t_l\}_{l=n+1}^{n+L}$ associated with all datasets such that time intervals $\{\Delta t_i\}_{i = 1}^{n+1}$ and $ \{\Delta t_l\}_{l=n+1}^{n+L}$ always stay between $0$ and $1$. 
Therefore, the bijective map $F_1$ will ensure that the outputs from the exponential probabilistic layer will stay in our desired range~\citep{dillon2017tensorflow}. 
Moreover, Figure~\ref{fig:Datahistograms} depicts that the time-interval histograms for all datasets follow an approximate exponential behavior, indicating that we do not require much flexibility in choosing the temporal bijective layer.

\subsection{Comparing Our Proposed Network with a Conventional Transformer Model}
\label{app: comparison}
For simplicity, in this section, we only talk about the spatial aspect of events, where everything is applicable to the events' temporal information.
The conventional seq2seq nature of the transformer architecture as a self-supervised learning process relies on using the ground-truth $\{{\mathbf{x}}_i\}_{i=1}^{n}$ only when minimizing the loss during training via a sequential $\{{\mathbf{x}}_l\}_{l=n+1}^{n+L}$ forecasting. 
In that case, $\{{\mathbf{x}}_l\}_{l=n+1}^{n+L}$ are \textbf{not} fed to the decoder block during training. 
More precisely, in the conventional transformer architecture, we use $\{{\mathbf{x}}_i\}_{i=1}^{n}$ to predict $\hat {\mathbf{x}}_{n+1}$ and then use the ground truth ${\mathbf{x}}_{n+1}$ to minimize the loss associated with predicting ${\mathbf{x}}_{n+1}$, where $\enspace \hat{} \enspace$ denotes the predicted value. 
Assuming that we have correctly predicted $\hat {\mathbf{x}}_{n+1}$, we use the ground-truth value ${\mathbf{x}}_{n+1}$ to continue with further predicting $\hat {\mathbf{x}}_{n+2}$. In the same way, we predict $\hat {\mathbf{x}}_{n+3},\hat {\mathbf{x}}_{n+4}, \cdots, \hat {\mathbf{x}}_{n+L}$ using the ground-truth values ${\mathbf{x}}_{n+2}, {\mathbf{x}}_{n+3}, \cdots, {\mathbf{x}}_{n+L-1}$, respectively.
However, during the testing phase, we aim to sequentially predict $\{\hat{ \mathbf{x}}_l\}_{l=n+1}^{n+L}$ as the decoder block's outputs by only relying on the input sequence $\{{\mathbf{x}}_i\}_{i=1}^{n}$ and (when predicting $\{\hat {\mathbf{x}}_l\}_{l>n+1}$) the respective predicted one-step-behind output values $\{\hat{\mathbf{x}}_l\}_{l=n+1}^{n+L-1}$. 

Nevertheless, the goal of our proposed network is to predict simultaneous distributions $\{\hat p_{l}^{({\mathbf{x}})}\}_{l=n+1}^{n+L}$ associated with events $\{{\mathbf{x}}_l\}_{l=n+1}^{n+L}$.
One potential way of achieving this is to never look at the ground truth values $\{{\mathbf{x}}_l\}_{l=n+1}^{n+L}$ while training (as in the conventional transformer architecture) and do the following process during training:

Use $\{{\mathbf{x}}_i\}_{i = 1}^{n}$ as inputs to the encoder and extract the hidden representations $\{\mathbf{h}_{{\mathbf{x}}_i}\}_{i = 1}^{n}$ as inputs to the decoder to output the first hidden representation $\mathbf{h}_{{\mathbf{x}}_{n+1}}$ associated with event ${\mathbf{x}}_{n+1}$. Input $\mathbf{h}_{{\mathbf{x}}_{n+1}}$ to the probabilistic layer, followed by the normalizing flow to output $\hat p_{n+1}^{({\mathbf{x}})}$ which is the predicted conditional spatial distribution associated with event ${\mathbf{x}}_{n+1}$. Next use the true sample ${\mathbf{x}}_{n+1}$ to minimize the loss associated with learning this distribution ($\hat p_{n+1}^{({\mathbf{x}})}$).
To model the conditional distribution $\hat p_{n+2}^{({\mathbf{x}})}$, use the same input values $\{{\mathbf{x}}_i\}_{i = 1}^{n}$ as well as the the ground-truth event ${\mathbf{x}}_{n+1}$ that is available to us during the training phase and is fed to the decoder. Continue this process until you reach $\hat p_{n+L}^{({\mathbf{x}})}$.

While the above-mentioned strategy is feasible and follows the conventional transformer network training,
in the test phase we first need to sample from the predicted $\hat p_{n+1}^{({\mathbf{x}})}$ to further predict $\hat p_{n+2}^{({\mathbf{x}})}$, then sample from $\hat p_{n+2}^{({\mathbf{x}})}$ to predict $\hat p_{n+3}^{({\mathbf{x}})}$, until we reach $\hat p_{n+L}^{({\mathbf{x}})}$. This strategy is highly prone to error accumulation as now we need the extracted samples $\{\hat {\mathbf{x}}_l\}_{l=n+1}^{n+L}$ to be perfectly close to the ground-truth values $\{{\mathbf{x}}_l\}_{l=n+1}^{n+L}$. 
To overcome this problem, while maintaining a seq2seq prediction, we proposed to have simultaneous/batched distribution learning that can be implemented in TensorFlow\footnote{\url{https://www.tensorflow.org/probability/api_docs/python/tfp}}. In this regard, we removed the look-ahead mask layer in the decoder block of the conventional transformer architecture to have the decoder output as a sequence of simultaneous events (hidden representations $\{\mathbf{h}_{{\mathbf{x}}_l}\}_{ = n+1}^{n+L}$).

To include the short-range dependencies among the output events $\{{\mathbf{x}}_l\}_{l=n+1}^{n+L}$ despite having a simultaneous prediction, we indeed use the information associated with these events in the training phase as the inputs to the decoder block. This leads to tuning the decoder layer weights and biases towards estimating the probability densities of several events in the future. However, in the test phase, we still obtain an estimation of the batched probability densities despite zero-padding the input to the decoder, meaning $\{{\mathbf{x}}_l\}_{l=n+1}^{n+L}=0$.
This happens due to the reliance of the network on the historical input data $\{{\mathbf{x}}_i\}_{i=1}^{n}$, as well as the learned bias associated with the decoder layer. To further assist the decoder to distinguish between different zero-padded inputs in the test phase, we used positional embedding (PE) proposed by \cite{vaswani2017} for both training and testing of our model.


\subsection{Data Normalization and Computing the Loss}
\label{app:datanorm}
We normalized all datasets after splitting the data into train, validation, and test sequences.
For location and the extra markers shown by $M$, we use the mean and variance of the training set to normalize the training, validation, and test sets. Therefore, applying our trained network to the unseen test sequences will result in predicting batched spatial distributions that are spanned in a normalized $2$- or $3$-dimensional grid as shown in Figures~\ref{fig:earthtransformer} and \ref{fig:earthbench}. We use the training set's mean and variance to denormalize the samples extracted from the high probability regions associated with the predicted spatial distributions. This way we compute the maximum log likelihood of regions in the normalized space. 

However, we took a different approach for time normalization as time, as opposed to other markers, has an increasing behaviour. More precisely, we use a similar sequence-time formation and normalization approach that was used in \cite{chen2020neural}.
In this case, before any data splitting, when we form sequences of size $n+L$ in length, we reformulate the times $t_{1}$ to $t_{n+L}$ in each sequence to approximately start from $0$. We then calculate the time intervals $\Delta t_{1}$ to $\Delta t_{n+L}$ associated with these sequences.

Next, we split the data into train, validation, and test input and output sequences, and normalize all $n$ input events to stay between 0 and 1. The same approach is taken for the $L$ output events. Therefore, all $n$ input and $L$ output events are normalized using the min and max of their own data sequence (meaning $0\leq \{t_{i}\}_{i = 1}^{n}\leq 1$ and $0\leq \{t_{l}\}_{l = n+1}^{n+L}\leq 1$). This step is needed for computational consistency before entering the time events $\{t_{i}\}_{i = 1}^{n}$ and $\{t_{l}\}_{l = n+1}^{n+L}$ into the encoder and decoder blocks, respectively. However, as we always convert the time units to satisfy $0\leq\{\Delta t_{i}\}_{i = 1}^{n}\leq 1$ and $0\leq\{\Delta t_{l}\}_{l = n+1}^{n+L}\leq 1$, no further normalization is needed for the time-intervals.

As shown in Figures~\ref{fig:earthtransformer}, \ref{fig:earthbench}, and \ref{fig:pinwheeltime}, in the test phase, we extract time-intervals $\Delta \hat t_{n+1}$ to $\Delta \hat t_{n+L}$ from the predicted batched temporal distributions to compute the occurrence time $ \hat t_{n+1}$ to $\hat t_{n+L}$ of multiple events in the future according to the following:

\begin{enumerate}
    \item Simultaneously extract the time-intervals $\{\Delta \hat t_{l}\}_{l=n+1}^{n+L}$ via sampling (mean of 1000 samples) from the predicted batched temporal distributions $\{\hat p_l^{(t)}\}_{l=n+1}^{n+L}$.
    \item To recover the predicted time values $\{ \hat t_{l}\}_{l=n+1}^{n+L}$, add up the extracted time-intervals $\{\Delta \hat t_{l}\}_{l=n+1}^{n+L}$ by sequentially computing $\hat t_{l} = \Delta \hat t_{l} +  \hat t_{l-1}$ for $\{l : n+1, \cdots, n+L\}$ where $\hat t_n = t_n$ is part of the input that is already given.
\end{enumerate}

As the final note, we mentioned earlier that as indicated by  (\ref{eq:sol_problem}) the predicted batched spatial distributions $\{\hat p_l^{({\mathbf{x}})}( \, \cdot \, ;t_l, \mathbf{h}_{{\mathbf{x}}_l})\}_{l=n+1}^{n+L}$ depend on the time event $\{t_l\}_{l = n+1}^{n+L}$. As illustrated in Figure~\ref{fig:network}, we designed our model to jointly predict both batched temporal and spatial distributions by first drawing a time sample $\hat t_l$ from $\hat p_l^{(t)}$ (according to what we described in the above paragraph) and then fixing and using $\hat t_l$ as an input to the batched spatial probabilistic layer to predict $\hat p_l^{({\mathbf{x}})}( \, \cdot \, ;\hat t_l, \mathbf{h}_{{\mathbf{x}}_l})$.
Since our baselines are designed to learn either temporal or spatial distributions, we decided to compare the baselines with our proposed method in a different way. For that matter, we simultaneously predict both $\hat p_l^{(t)}( \, \cdot \, ;h_{t_l}) $ and $\hat p_l^{({\mathbf{x}})}( \, \cdot \, ;t_l, \mathbf{h}_{{\mathbf{x}}_l})$, where for $\hat p_l^{({\mathbf{x}})}( \, \cdot \, ;t_l, \mathbf{h}_{{\mathbf{x}}_l})$ we assume that time event $t_l$ is known and given in advance. 
Therefore, the loss error is comparable with the baselines that predict solely spatial distributions, since we are controlling for the error that could be introduced by a wrong time prediction $\hat t_l$.

\subsection{Modeling Batched Distributions Associated with the Extra Marker $M$}

As shown in Figure~\ref{fig:network}, during training, both batches of inputs $\{(t_i, {\mathbf{x}}_i,M_i)\}_{i = 1}^n$ and outputs $\{(t_l, {\mathbf{x}}_l,M_l)\}_{l = n+1}^{n+L}$ are fed into the encoder and the decoder, respectively. 
Using (\ref{eq:losstransformer}), the goal is to learn a parametric description of a batched probability distribution that seeks to explain future events in only time and space, meaning $p_l^{(t)}$ and $p_l^{({\mathbf{x}})}$, respectively. 
The reason behind considering only these markers (time and space) in this work is that these are the main markers covered by the original Hawkes process as shown in (\ref{eq:lambda}). As indicated by the Hawkes intensity function in (\ref{eq:lambda}), using other markers $\{M_i\}_{i = 1}^{n}$ as historical information has no contradiction with learning only the batched spatio-temporal distributions.
However, the same principle as the one proposed in this work could be used to learn/predict marker distributions $p_l^{(M)}$ associated with multiple events in the future. In specific cases such as when marker $M$ represents the earthquake magnitude in earth science, it might be useful to also consider the Gutenberg-Richter (GR) frequency-magnitude law~\citep{gutenberg1944frequency, knopoff2000magnitude, tinti1985effects, rhoades1996estimation}. 

\section{Ablation Study}
\label{app:ablation}
To further leverage the importance of each layer/block we used in our proposed model, in addition to what we described in Sections~\ref{sec:network} and~\ref{sec:limitations}, we also performed two ablation studies on our network using all the available datasets.

\begin{table*}[th]
\caption{Results representing the negative log-likelihood for the output events in the range $n+1$ to $n+L$ for the learned distributions including ablation studies (less is better). $\pm$ indicates the standard deviation of the loss among all test batch sequences.}
\label{table:ablation}
\vskip 0.15in
\begin{center}
\begin{small}
\begin{sc}
\begin{tabular}{lcccr}
\toprule
& Earthquake & Citibike &Covid-19& Pinwheel \\
\midrule
Our network ($p_l^{(t)}$)   & $0.33_{\pm 0.17}$ & $-6.279_{\pm  0.123}$ & $-5.87_{\pm  0.09}$ & $1.27_{\pm 0.26}$\\
Our network ($p_l^{({\mathbf{x}})}$)   & $2.03_{\pm 0.26}$ & $-3.28_{\pm 0.14}$ & $2.18_{\pm 0.21}$ & $1.57_{\pm 0.09}$\\
Our network ($p_l^{(t, {\mathbf{x}})}$)   & $2.36_{\pm 0.36}$ & $-9.56_{\pm 0.18}$ & $-3.7_{\pm 0.16}$ & $2.84_{\pm 0.19}$ 
\\ \hline \\
History independent ($p_l^{(t)}$)   & $1.26_{\pm 0.26}$ & $-6.3_{\pm  0.15}$ & $-5.9_{\pm  0.22}$ & $1.27_{\pm 0.27}$\\
History independent ($p_l^{({\mathbf{x}})}$)   & $2.49_{\pm 0.32}$ & $-3.27_{\pm 0.08}$ & $1.64_{\pm 0.05}$ & $7.06_{\pm 0.03}$\\
History independent 
($p_l^{(t, {\mathbf{x}})}$)   & $3.75_{\pm 0.46}$ & $-9.58_{\pm 0.14}$ & $-4.26_{\pm 0.09}$ & $8.334_{\pm 0.26}$ 
\\ \hline \\
No decoder ($p_l^{(t)}$)   & $0.37_{\pm 0.18}$ & 
$-6.076_{\pm 0.13}$ & $-5.41_{\pm  0.12}$ & $1.29_{\pm 0.25}$\\
No decoder ($p_l^{({\mathbf{x}})}$)   & $2.21_{\pm 0.3}$ & $-3.1_{\pm  0.09}$ & $2.2_{\pm 0.32}$ & $0.61_{\pm 0.06}$\\
No decoder ($p_l^{(t, {\mathbf{x}})}$)   & $2.58_{\pm 0.41}$ & $-9.17_{\pm 0.14}$ & $-3.21_{\pm 0.28}$ & $1.9_{\pm 0.18}$ \\
\bottomrule
\end{tabular}
\end{sc}
\end{small}
\end{center}
\vskip -0.1in
\end{table*}

\subsection{History Independency}
\label{app:histindept}
In this study, we assume that $p_l^{({\mathbf{x}})}$ and $p_l^{(t)}$ are completely independent of the history, meaning that we have $p_l^{({\mathbf{x}})}( \, \cdot \, ; t_l, \mathbf{h}_{{\mathbf{x}}_l}) = p_l^{({\mathbf{x}})}( \, \cdot \, ; t_l)$ and $p_l^{(t)}( \, \cdot \, ;h_{t_l}) =  p_l^{(t)}$. 
For this matter we provide batches of zero inputs to the encoder $\{(t_i, {\mathbf{x}}_i,M_i)\}_{i = 1}^n = 0$ in both the training and testing phase while keeping all other steps exactly as before.
Results for this study on different datasets are given in Table~\ref{table:ablation}. 

\textbf{South California Earthquakes:} As shown by Table~\ref{table:ablation}, removing history dependency from our network causes worse performance of multi-event forecasting in both time and space for this dataset. 
This is in line with the well-known epidemic-type-aftershock-sequence (ETAS) model~\citep{ogata1998space} suggesting long-term and long-range dependencies in both time and space between seismic events.

\textbf{Citibike:} Removing history dependency on this data did not highly affect the results.
Our interpretation of this dataset is that many other factors other than the previous bikers' markers included in this dataset might influence the time or location of riding a bike. 
Therefore, removing history dependency does not highly influence forecasting the time or location of multiple bikers in the future when no other influencing factors are considered.

\textbf{Covid-19:} We realized that removing the history dependency did not affect time distribution forecasting, but instead enhanced predicting space distributions. 
Despite the disease being very contagious, the space distribution forecasting results are not surprising at all.
This is due to the fact that we are taking into account the long-range history dependency over many different states of the United States. 
Unless we consider trips between states, the patients' time and location in one state might even provide false information influencing the chances of others getting the disease in another state. 
Therefore, removing this information might even enhance predicting space distributions.
Our interpretation of very little changes in forecasting time distribution in the case of having no historical information is that covid-19 is among the fastest disease to catch, therefore not very long-term dependency in time is needed to forecast the time of catching the disease. 
We guess that the short-term dependency captured by the decoder during training is enough to predict when the next person will catch the disease.

\textbf{Pinwheel:} As shown on Figure~\ref{fig:pinwheel} top, the sequences of pinwheel events are formed in a clock-wise way. Therefore, historical positional and temporal information should have a big affect on predicting the location of next events in the future. 
This is in line with the results shown in Table~\ref{table:ablation}, indicating how removing historical information worsens spatial multi-event distribution forecasting.
On the other hand, despite using synthetic Hawkes time~\citep{RHawkes} associated with pinwheel events, we saw no change in time forecasting when removing history dependency. 
For this outcome, we infer that the thining process~\citep{ogata1993fast} used to extract the times~\citep{RHawkes} assigned to the events does not consider long-term and long-range dependency. 
Therefore, the short-term history dependency that exists among the output events $\{(t_l, {\mathbf{x}}_l)\}_{l = n+1}^{n+L}$ that is captured during training by the decoder layer, suffices for predicting multiple events' time distributions with no reliance on further look onto the history.

\subsection{No Decoder}
\label{app:nodecoder}
Here, we no longer feed $\{(t_l, {\mathbf{x}}_l,M_l)\}_{l = n+1}^{n+L}$ to the decoder during the training phase. This is similar to the test phase of our model where we input $\{(t_l, {\mathbf{x}}_l,M_l)\}_{l = n+1}^{n+L} = 0$ to the already trained network. However, in this study we input $\{(t_l, {\mathbf{x}}_l,M_l)\}_{l = n+1}^{n+L} = 0$ during both training and testing phases. 
Results in Table~\ref{table:ablation} indicate that except for the pinwheels dataset's spatial distribution forecasting, removing the decoder worsens the prediction results.
As earlier mentioned in Section~\ref{sec:limitations} during training, the network learns to fit the distribution of multiple future events jointly, in a self-supervised setting.
Hence, during testing, zero-padding the input to the decoder still obtains an estimation of the spatio-temporal probability densities of several events in the future. 
This happens due to the reliance of the network on the historical input data, as well as the learned bias associated with the decoder layer. 
However, we accept that the learned bias might only provide little history dependency information between output events, especially in the case of a small number of events to be predicted ($L = 3$ in our case). 
Since the setup when having no decoder differs from the main setup of our proposed network only in terms of the learned bias associated with the decoder layer, we do not expect to see much difference in the results.

In the case of the pinwheel dataset, we realized that removing the decoder enhanced the multi-event spatial distribution forecasting results. 
This is when the batched temporal distribution forecasting results are negatively affected. 
In line with the outcomes of the previous ablation study, we inferred that the pinwheel events are influenced massively by the long-term  history dependency in space.
Therefore, removing the short-term dependency among the output events will allow the events to focus more on long-range effects of the historical data. However, the results associated with predicting the temporal distributions got worse. As expected from the outcome of the study in Appendix~\ref{app:histindept}, this might be due to the importance of short-term dependency between the time of the events, modeled by the thining process~\citep{ogata1993fast}.

\subsection{Conclusion on Ablation Study}
\label{app:ablationconlusion}
Through these two studies and what we discussed in the main body of this paper, we showed the critical role of each layer forming our network. 
Mainly, we observed that each layer is designed for a specific task, driven by the physics behind the formation of the used dataset. 
Unlike other methods that solely rely on the Hawkes process, our network is designed to fully consider any sort of short- or long-term history dependencies imposed by the physics behind data. 

In general, from these studies, we concluded that the performance of our proposed network is highly data-driven. 
Predicting batched distributions associated with datasets that are formed based on both long- and short-range history dependencies, such as the earthquake dataset, demands the usage of both the encoder and decoder blocks of our model. 
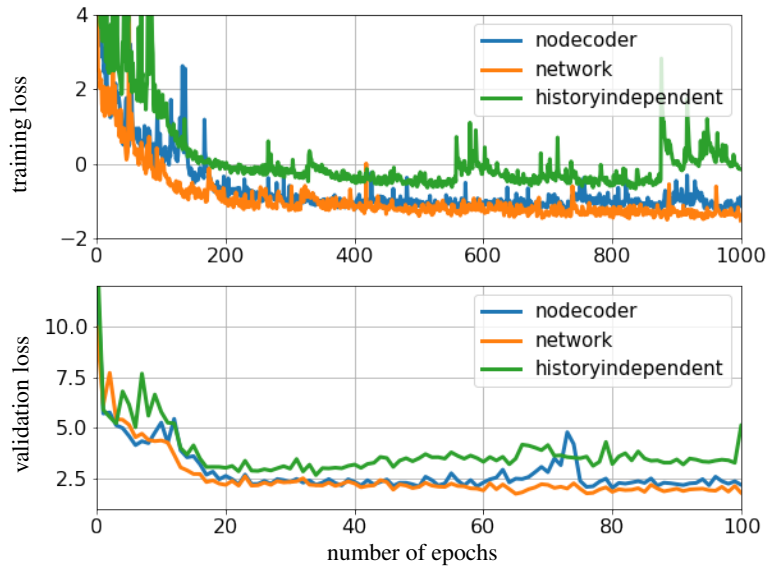
\begin{figure}[h!]
\begin{minipage}{0.4\linewidth}

        When events are more influenced by short-term dependencies, or there's a lack of the presence of other factors influencing the events' occurrence, the encoder might negatively affect the prediction results. 
        This was shown as the outcome of the first study on the Covid-19 dataset. 
        The reverse applies to events dependent on long-range history dependencies, such as the pinwheel and earthquake datasets.
        There might be cases where removing the encoder has no influence on the prediction results, due to no significant history dependency, such as the Citibike dataset.

        On the other hand, capturing the short-term dependencies, mainly done by the decoder layer, is important when events are also influenced by these dependencies. 
        This conclusion is based on the outcomes of applying the second ablation study to many datasets, such as the earthquake dataset, Citibike, Covid-19, and the temporal events forming the Pinwheel dataset.

        Since all datasets used in this work, except for the pinwheel dataset, represent real-world phenomena, we believe no single model can be used to learn/predict distributions associated with them.
        Therefore, it is critical to have a network that can be driven by the data's underlying physics, whilst being capable of simultaneous multi-event prediction.
    \end{minipage}\hspace{0.2cm}
\begin{minipage}{0.6\linewidth}
    \begin{tikzpicture}
    \node (img)  {\includegraphics[width=0.95\linewidth]{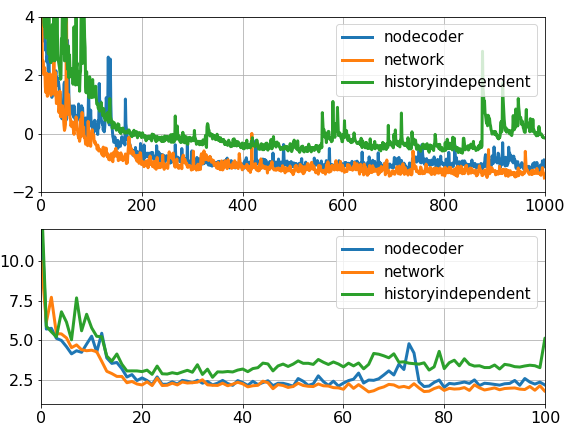}};
    \node[left=of img, node distance=0cm, rotate=90, anchor=center,yshift=-0.8cm,xshift=1.8cm,font=\color{black}] {\footnotesize training loss};
    \node[left=of img, node distance=0cm, rotate=90, anchor=center,yshift=-0.8cm,xshift=-1.8cm,font=\color{black}] {\footnotesize validation loss};
    \node[below=of img, node distance=0cm,  anchor=center,yshift=1.0cm,xshift=0cm,font=\color{black}] {\footnotesize number of epochs};
    \end{tikzpicture}
    \caption{Comparing the performance of the proposed network with two ablation studies applied to the earthquake data. 
    Both studies worsen the prediction results.
    This highlights the importance of both long- and short-range history dependencies among seismic events.}
    \end{minipage}
\end{figure}

\section{More Results and Analysis}
\label{app:analysis}

\begin{figure}[ht]
\begin{minipage}{\linewidth}
    \begin{minipage}{1.0\linewidth}
    \begin{tikzpicture}
    \node (img)  {\includegraphics[width=0.95\linewidth]{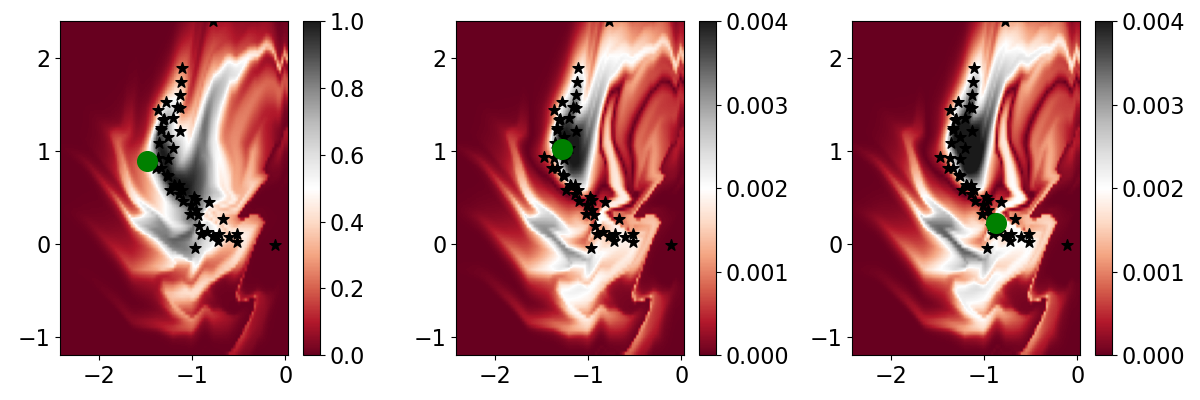}};
    \node[left=of img, node distance=0cm, rotate=90, anchor=center,yshift=-0.6cm,xshift=0cm,font=\color{black}] {\footnotesize Our Network};
    \end{tikzpicture}
    \end{minipage}
    \vspace{-0.1cm}
    \\
    \centering
    \begin{minipage}{1.0\linewidth}
    \begin{tikzpicture}
    \node (img)  {\includegraphics[width=0.95\linewidth]{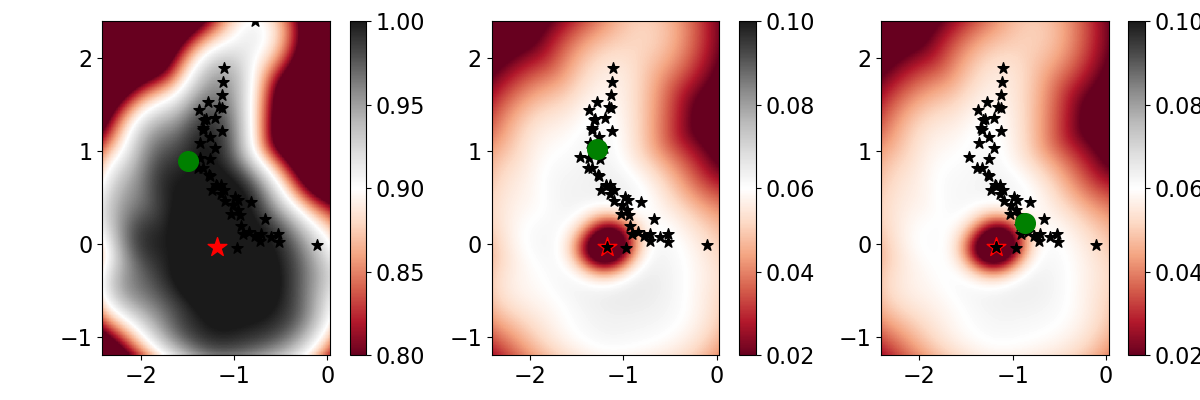}};
    \node[left=of img, node distance=0cm, rotate=90, anchor=center,yshift=-1.5cm,xshift=0cm,font=\color{black}] {\footnotesize conditional GMM};
\end{tikzpicture}
\end{minipage}
\end{minipage}
\hspace{0.1cm}
\\
    \begin{minipage}{\linewidth}
        \caption{Visualizations of the learned history dependent batched spatial distributions for the last three events of simulated Hawkes pinwheel sequence using \textbf{our proposed architecture} vs. \textbf{conditional GMM}. The first figure on the left hand side depicts the predicted distribution associated with the first event, where the middle and the right hand side figures only depict the consecutive differences between the predicted densities. The green circle shows the true event that is expected to occur. The black stars represent the last 50 events of the input sequence as history information. The red star on bottom figures shows the sequentially predicted next event using (\ref{eq: firstmoment}) which needs to be injected back to the history for the next step prediction. Due to high error accumulation, we instead use the green circle to update history on the bottom figures.}
\label{fig:pinwheeltransformer}
    \end{minipage}
\end{figure}

To show the performance of our network on synthetic Hawkes time simulated data vs. the baseline model, we used the pinwheel dataset as introduced in Section~\ref{sec: data}, where the events are distributed with less bias towards a specific region. As depicted in Figure~\ref{fig:pinwheeltransformer} on top, not only the history dependent batched spatial distributions for the next three events are well predicted using our network, but also regions prone to events occurring further away in time, shown in a clock-wise pattern in Figure~\ref{fig:pinwheel}, are as well predicted. On the other hand, the conditional GMM model was not successful in extracting the complex distribution associated with next events' occurrences. Note that using the conditional GMM model to perform sequential next event prediction, at each step, we incorporate the true event shown by the green circle, rather than the newly predicted event shown by the red star, to redefine the history for next step prediction. This is due to the high error accumulation caused by using the sequentially predicted events. 

\begin{figure}[ht]
\begin{minipage}{\linewidth}
    \begin{minipage}{1.0\linewidth}
    \begin{tikzpicture}
    \node (img)  {\includegraphics[width=0.95\linewidth]{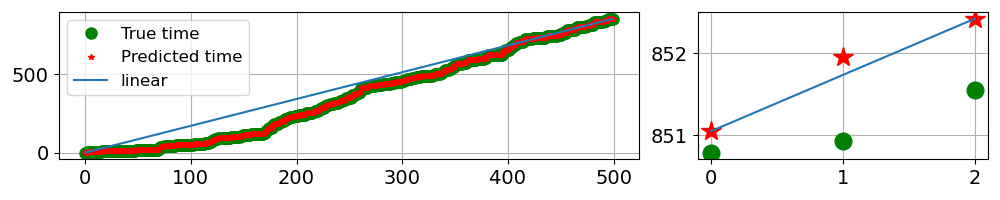}};
    \node[left=of img, node distance=0cm, rotate=90, anchor=center,yshift=-0.8cm,xshift=0cm,font=\color{black}] {\footnotesize Our Network};
    \end{tikzpicture}
    \end{minipage}
    \vspace{-0.1cm}
    \\
    \centering
    \begin{minipage}{1.0\linewidth}
    \begin{tikzpicture}
    \node (img)  {\includegraphics[width=0.95\linewidth]{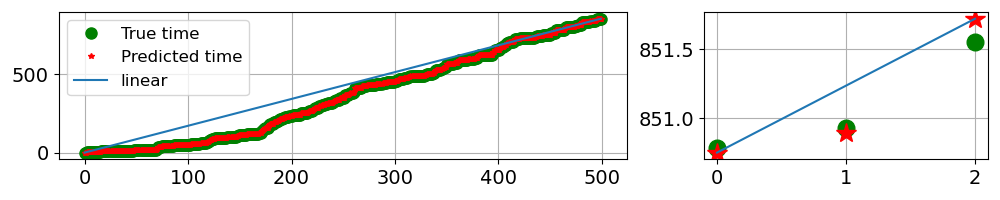}};
    \node[left=of img, node distance=0cm, rotate=90, anchor=center,yshift=-0.8cm,xshift=0cm,font=\color{black}] {\footnotesize Hawkes model};
\end{tikzpicture}
\end{minipage}
\end{minipage}
\hspace{0.1cm}
\\
    \begin{minipage}{\linewidth}
        \caption{The true (green circles) vs. predicted (red stars) time of the last three events of the simulated Hawkes pinwheel sequence using {\bf our model} on top and the {\bf time Hawkes model} on the bottom. The results on top are generated simultaneously, whereas the bottom results are predicted sequentially via updating history based on the most recently predicted event time. The linear line is indicative of predicting same time-intervals between events.}
\label{fig:pinwheeltime}
    \end{minipage}
\end{figure}

Figure~\ref{fig:pinwheeltime} shows the predicted time of the next three events applying {\bf our proposed network} \textit{and} the {\bf time Hawkes model} to the simulated Hawkes pinwheel dataset. While the result on the top figure are indicative of simultaneous multiple future time event prediction, the bottom figures depict utilizing (\ref{eq: firstmoment}) sequentially while updating the history via the most recently predicted event. These results are in agreement with Table~\ref{table:result} indicating the outperformance of using the Hawkes model for this dataset. As mentioned earlier, this is due to the fact that the time assigned to the synthetic Hawkes dataset are simulated by performing thinning algorithm on a time Hawkes process. As mentioned earlier, the expected times extracted by applying our network to this and all other datasets are measured by taking the average of 1000 samples taken from the batched history dependent predicted time distributions.

\begin{figure}[h]
    \centering
    \includegraphics[width=0.94\textwidth]{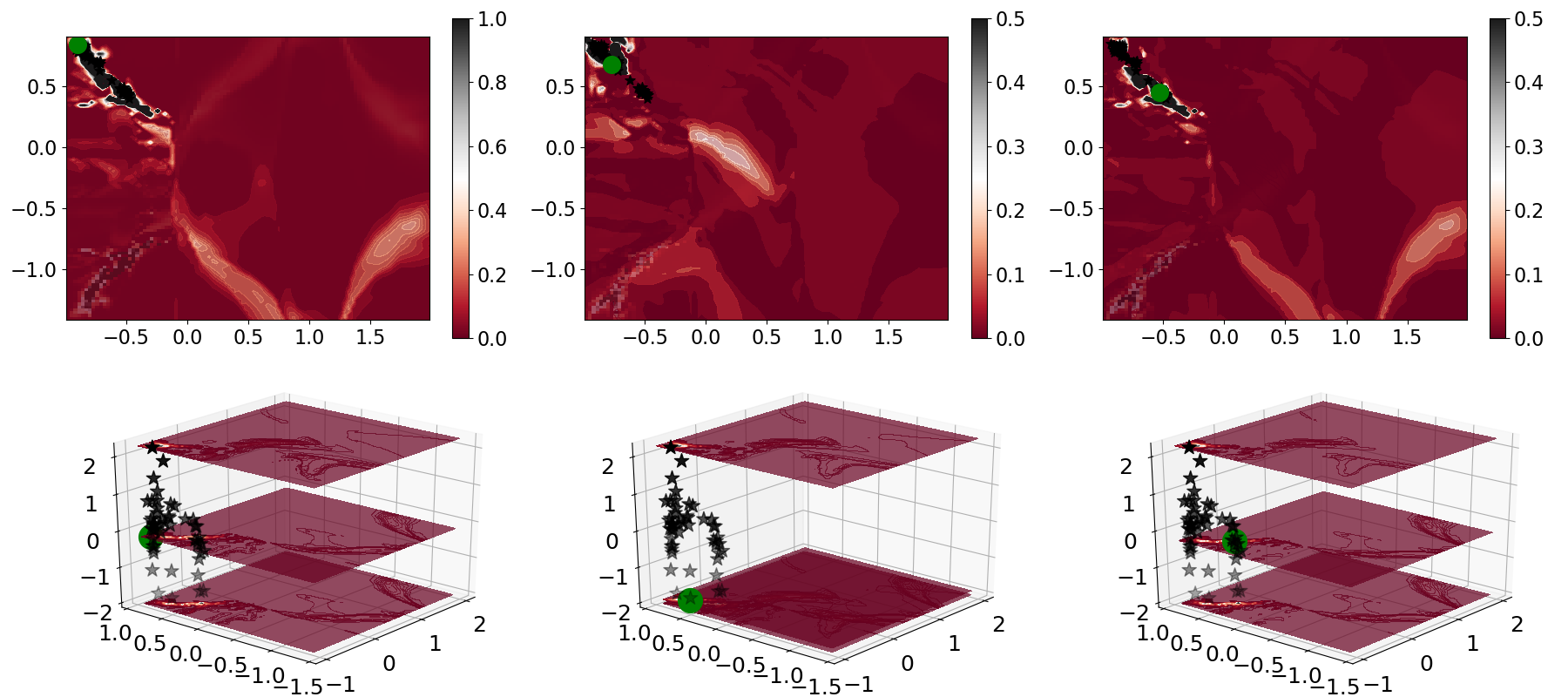}
    \caption{Visualizations of the learned history dependent batched spatial distributions for the last three events of an unseen earthquake sequence using \textbf{our proposed architecture}. The \textbf{bottom} figures depict the predicted batched spatial distribution associated with the last three events in 3 dimensions where the z axis is along events' depth. The 2-d representation of the last three events' predicted spatial distributions mapped to the events' depth are shown on the \textbf{top} plots where except for the one on the right, we show the differences in the predicted distributions. The black stars on the spatial distributions represent the last 50 events of the input sequence as the history information, where in this figure the history events are very much aligned along the main fault in South California. }
    \label{fig:earthtransformerextra1}
\end{figure}

\begin{figure}[h]
    \centering
    \includegraphics[width=0.94\textwidth]{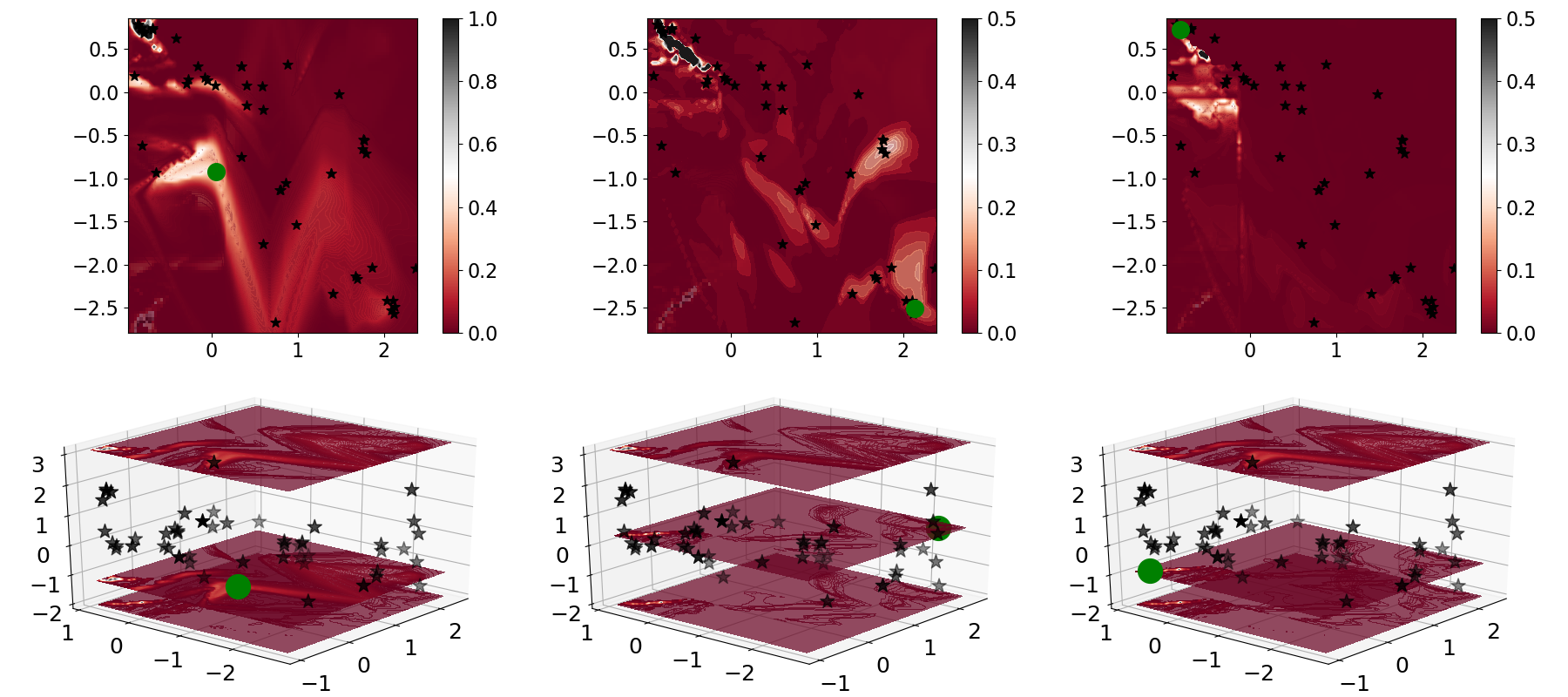}
    \caption{Same visualization as in Figure~\ref{fig:earthtransformerextra1} on another earthquake sequence of length 500 where the history events, shown by black stars, are more distributed along all fault lines rather than just one.}
    \label{fig:earthtransformerextra2}
\end{figure}
\begin{figure}[h]
\centering
    \begin{minipage}{0.8\linewidth}
    \begin{tikzpicture}
    \node (img)  {\includegraphics[width=1\linewidth]{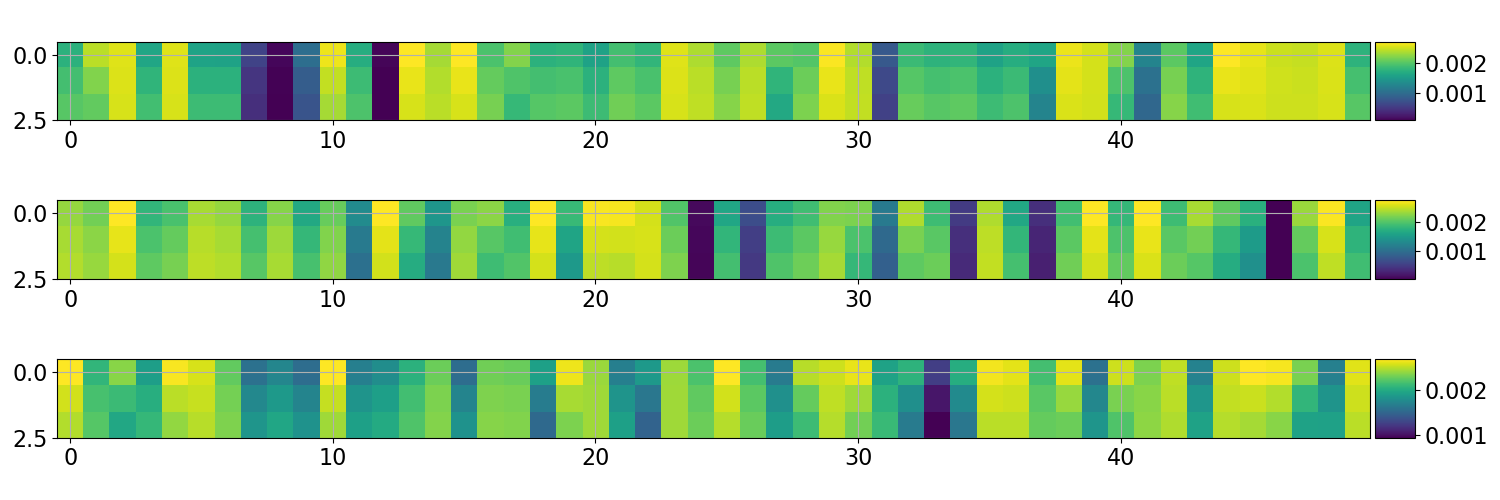}};
    \node[above=of img, node distance=0cm, yshift=-1.5cm,xshift = 0.0cm,font=\color{black}] {Encoder-decoder Layer 1};
    \node[left=of img, node distance=0cm, rotate=90, anchor=center,yshift=-0.8cm,xshift=0cm,font=\color{black}] {\footnotesize Attention heads 1-3};
    \end{tikzpicture}
    \end{minipage}
    \vspace{0.5cm}
    \\
    \centering
    \begin{minipage}{0.8\linewidth}
    \begin{tikzpicture}
    \node (img)  {\includegraphics[width=1\linewidth]{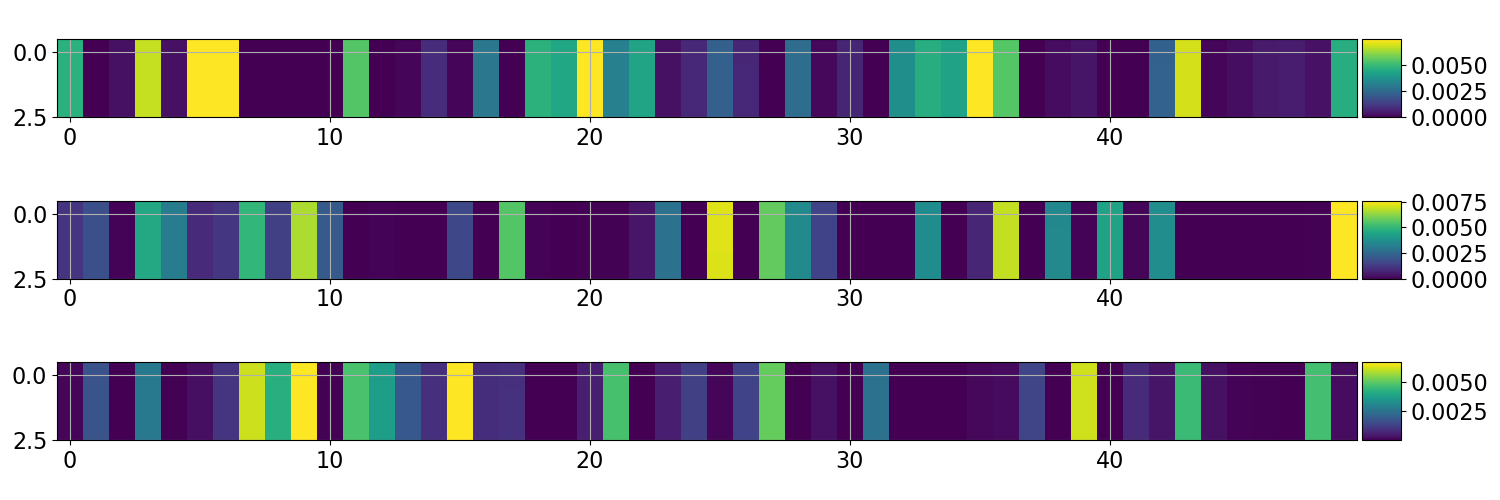}};
    \node[above=of img, node distance=0cm, yshift=-1.5cm,xshift = 0.0cm,font=\color{black}] {Encoder-decoder Layer 2};
    \node[left=of img, node distance=0cm, rotate=90, anchor=center,yshift=-0.8cm,xshift=0cm,font=\color{black}] {\footnotesize Attention heads 1-3};
\end{tikzpicture}
\end{minipage}
\vspace{0.5cm}
\\
    \centering
    \begin{minipage}{0.8\linewidth}
    \begin{tikzpicture}
    \node (img)  {\includegraphics[width=1\linewidth]{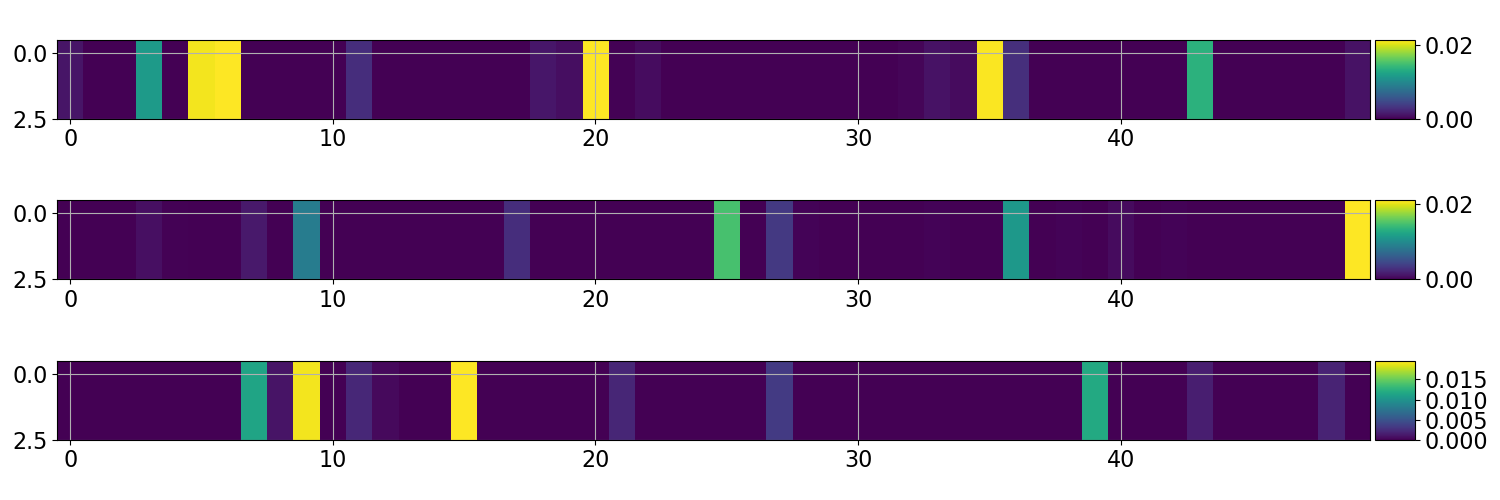}};
    \node[above=of img, node distance=0cm, yshift=-1.5cm,xshift = 0.0cm,font=\color{black}] {Encoder-decoder Layer 3};
    \node[left=of img, node distance=0cm, rotate=90, anchor=center,yshift=-0.8cm,xshift=0cm,font=\color{black}] {\footnotesize Attention heads 1-3};
\end{tikzpicture}
\end{minipage}
\hspace{0.4cm}
\centering
    \\
    \begin{minipage}{1\linewidth}
        \caption{Visualization of the first 3 out of 6 encoder-decoder attention layers, where we show 3 out of 6 heads per layer. These figures are the results of applying \textbf{our proposed network} to the pinwheel dataset. The x axis represent historical events, whereas the y axis show the three expected events in the future. The colors show the intensity of the learned scores representing how the past events impact the occurence of future events.}
\label{fig:attentionscores}
    \end{minipage}
\end{figure}

\section{Baselines}

\label{app:baseline}
We use baselines to separately model the time and spatial events distributions, conditioned on the history. Earlier in (\ref{eq:loss}), we showed that the loss defined for the benchmarks have some differences with the loss we defined for our proposed architecture in the training phase. As already discussed in the main body of the paper, the baseline models only provide the expected next event using (\ref{eq: firstmoment}), which unlike our proposed method, gives the predicted event explicitly rather than a distribution. For further prediction in the future, the already predicted event has to be incorporated in the history, which will accordingly update the intensity function $\lambda(t_i|H_{t_i})$ as well as the space distribution $p({\mathbf{x}}_i|t_i, H_{t_i})$ to be used for further predictions. To have a precise updated history, we need our predicted points to be as close to the true values as possible. Therefore, we expand (\ref{eq:loss}) to provide further detail as
\begin{align}
\label{eq:loss_benchmark}
    \text{loss} &=\underbrace{-\sum_{i=1}^{n}\log p_t(t_i|H_{t_i}) -\sum_{i=1}^{n}\log p_{\mathbf{x}}({\mathbf{x}}_i|t_i, H_{t_i})}_{\text{negative log-likelihood encoding history data}} \\ \nonumber  &\underbrace{-\sum_{l=n+1}^{n+L}\log p_t(t_l| \Tilde{H}_{t_l}) -\sum_{l=n+1}^{n+L}\log p_{\mathbf{x}}({\mathbf{x}}_l|t_l, \Tilde{H}_{t_l})}_{\text{negative log-likelihood of predicted events conditioned on the updated history }} \\ \nonumber
    &\underbrace{+ \lambda_1\sum_{l=n+1}^{n+L}|t_l-\hat{t}_l| + \lambda_2\sum_{l=n+1}^{n+L}||{\mathbf{x}}_l-\hat{{\mathbf{x}}}_l||_2}_{\text{restricting the predicted values to match the true values}} ,
\end{align}
where $\Tilde{H}_{t_l}$ is the updated history including the most recent predicted event $\{\hat{t}_l, \hat{{\mathbf{x}}}_l\}$. As mentioned in Section~\ref{sec: training}, during the testing phase, we remove the first and third parts in (\ref{eq:loss_benchmark}) when computing the prediction error using the baseline models. This will cause a fair comparison between all prediction results provided in Table~\ref{table:result}.

\textbf{Conditional Gaussian mixture model}:
We implemented two forms for the Conditional Gaussian mixture models as the baselines to learn the spatial distributions $p({\mathbf{x}}_i|t_i, H_{t_i})_{i = 1:n}$ mentioned in Section~\ref{sec:baselines}. In the first method we compute the pairwise Gaussian log-likelihood among all ${\mathbf{x}}_i$ and  ${\mathbf{x}}_j$ associated with events  $1\leq j \leq i \leq n$, as well as the pairwise log-timedecay $\Delta t_{ij}$. We minimize the loss associated with adding up the pairwise Gaussian log-likelihood and log-timedecay to learn the history dependent Gaussian mixture model.

The second method is the general K-cluster Gaussian mixture model introduced in \cite{murphy2012machine, bishop2006pattern}, where the learned parameters are functions of the historical events markers. Having a mixture of K Gaussian densities 

\begin{align}
\label{eq:k_cluster}
    p({\mathbf{x}_l}|\mu, \Sigma) =\underset{K}{\Pi} \frac{1}{(2\pi)^{\frac{n_k}{2}}|\Sigma_k|^{\frac{1}{2}}}\text{exp}(-\frac{1}{2}({\mathbf{x}}_{l_k} - \mu_k)^{\top}\Sigma_k^{-1}({\mathbf{x}}_{l_k} - \mu_k)), \enspace 1\leq k \leq K
\end{align}
where $n_k$ represents the number of samples associated with cluster k, we learn $\mu_k$ and $\Sigma_k$ from  $\{\{(t_i, {\mathbf{x}}_i,M_i)\}_{i = 1}^n\}_k$ as the time, space and the extra marker associated with all events assigned to cluster k. In this case, $p({\mathbf{x}_l}|\mu, \Sigma)$ in (\ref{eq:k_cluster}), denotes the conditional spatial probability associated with event ${\mathbf{x}}_{l = n +1}$.

Although the second method is capable of conditionally being dependent on any other markers associated with events, the first method had a better performance on all datasets. Therefore, in this work we only reported the results based on the first method, where both implementations are given in our publicly available code.
\section{Preprocessing of Data Sequences}
\label{app:preprocessing}
\begin{figure}[ht]
\centering
\hspace{-1.0cm}
    \begin{minipage}{0.92\linewidth}
    \begin{tikzpicture}
    \node (img)  {\includegraphics[width=0.95\linewidth]{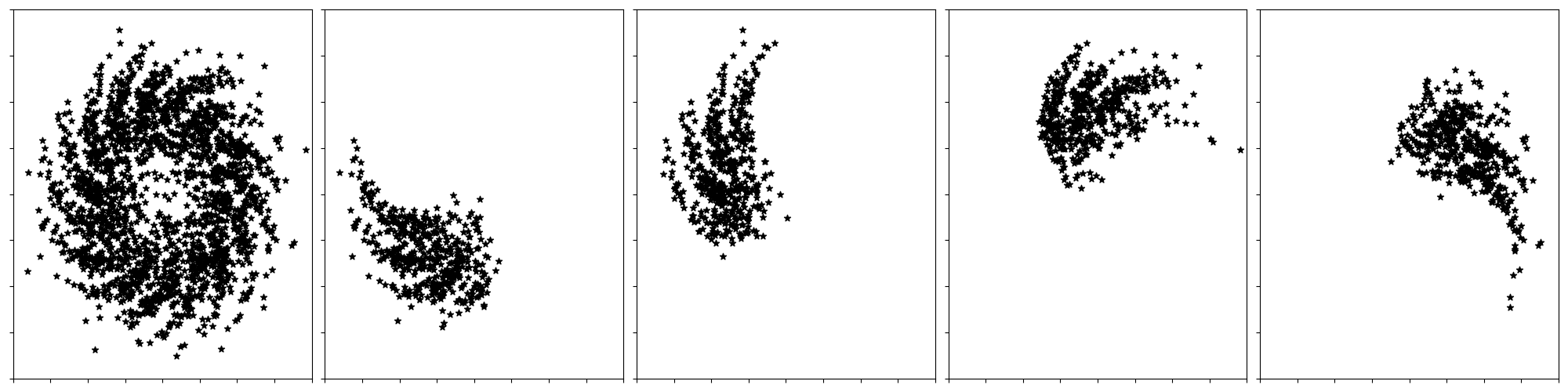}};
    \node[left=of img, node distance=0cm, rotate=90, anchor=center,yshift=-0.7cm,xshift=0cm,font=\color{black}] {\footnotesize Pinwheel};
    \end{tikzpicture}
    \end{minipage}
    \vspace{-0.22cm}
    \\
    \begin{minipage}{0.99\linewidth}
    \begin{tikzpicture}
    \node (img)  {\includegraphics[width=0.95\linewidth]{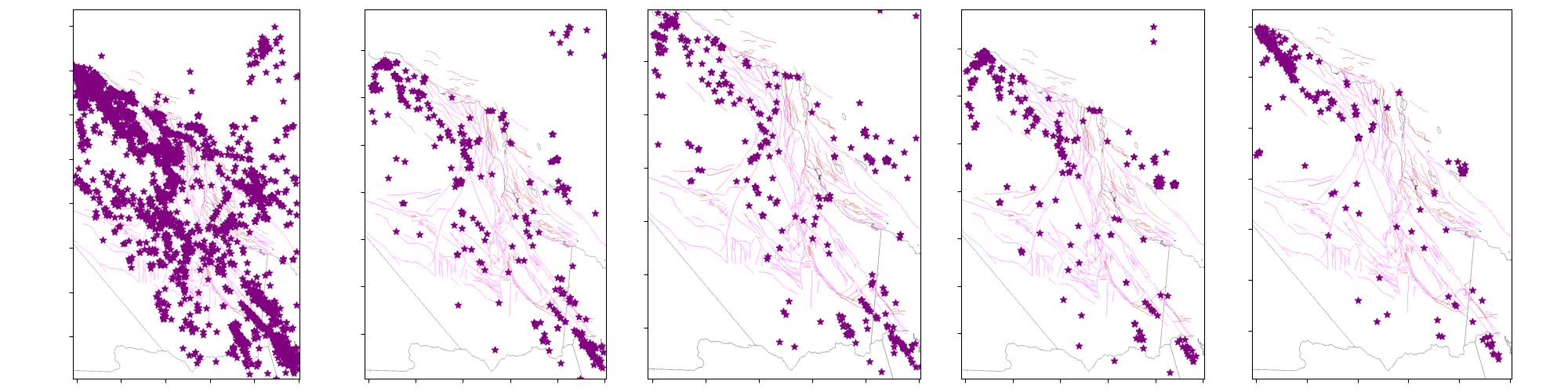}};
    \node[left=of img, node distance=0cm, rotate=90, anchor=center,yshift=-1.3cm,xshift=0cm,font=\color{black}] {\footnotesize Earthquake};
\end{tikzpicture}
\end{minipage}
\centering
    \\
    \begin{minipage}{1\linewidth}
        \caption{{\bf Top}: Pinwheel synthetic dataset where simulated Hawkes times are assigned to the events of clusters in a clockwise way. These timings are simulated using thinning algorithm \cite{ogata1981lewis}. {\bf Bottom}: South California earthquakes distributed over the main faults. {\bf Both}: All data points vs. consecutive sequences of 500 events with no overlap.}
\label{fig:pinwheel}
    \end{minipage}
\end{figure}

\textbf{South California Earthquakes.} We use earthquake discrete events distributed over South of California with latitude, longitude and depth. The distribution of all events along the main faults as well as four sequences of 500-length size, projected on a 2-d map of South California faults, are shown in Figure~\ref{fig:pinwheel}. Since we don't work with in-between distances, there is no need to convert to the Cartesian coordinate system. 
\begin{figure}[h!]
    \centering
    \includegraphics[width=\textwidth]{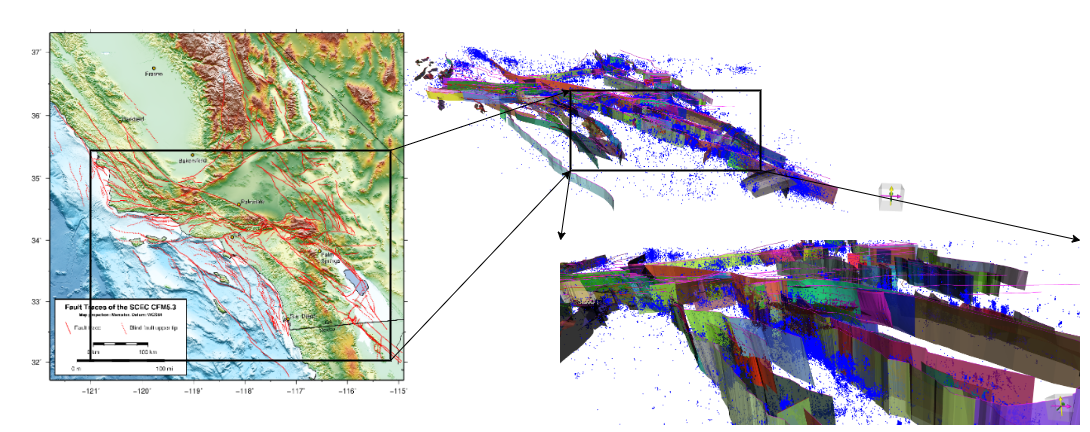}
    \caption{Visualization of distribution of discrete earthquake events on the South California main faults covering events from 2008 to 2016.}
    \label{fig:southcalifearth}
\end{figure}
After forming sequences as mentioned in Section~\ref{sec: data}, we normalize the space and magnitude of each sequence with respect to the mean and standard deviation gathered from all events. The bias caused by normalization towards more earthquake prone regions as previously shown in Figure~\ref{fig:earthtransformer} is remained to be eased in later work as discussed in Section \ref{sec:conclusion}.

\textbf{Citibike.} We use the data from April to August of 2019. No preprocessing was performed for this dataset. This resulted in 3500 training sequences, 900 validation sequences, and 600 test sequences of same length.

\textbf{Covid-19.} We use the data from March of 2020 to May of 2022. No preprocessing was performed for this dataset. This resulted in 1400 training sequences, 360 validation sequences, and 240 test sequences of same length.

\textbf{Pinwheel.} We sample from a multivariate process with 15 clusters, each containing 150 events. Due to the space Hawkes synthetic dataset not being available, we used the same spatial distribution provided in \cite{chen2020neural} with the assigned Hawkes timings using the RHawkes package \cite{RHawkes}. Since we have sequences of 500 events, each sequence will contain a few number of clusters as shown in Figure~\ref{fig:pinwheel}.  This resulted in 613 training sequences, 157 validation sequences, and 106 test sequences of same length.

\begin{figure}[h]
\centering
\begin{minipage}{1\linewidth}
    \begin{tikzpicture}
    \node (img)  {\includegraphics[width=1\linewidth]{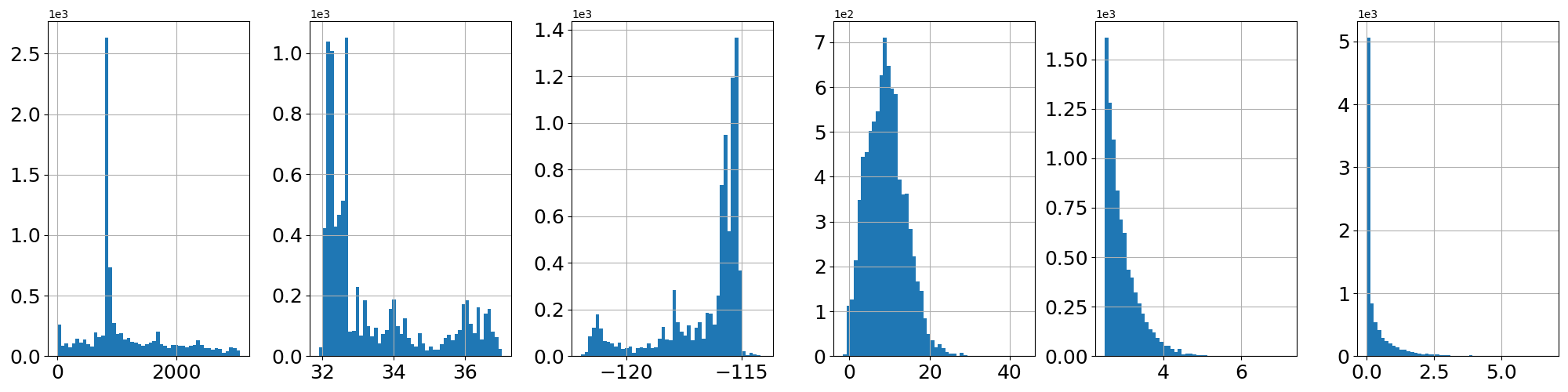}};
    \node[below=of img, node distance=0cm, yshift=1.2cm,xshift = -6.8cm,font=\color{black}] {\footnotesize Time};
    \node[below=of img, node distance=0cm, yshift=1.2cm,xshift = -4.0cm,font=\color{black}] {\footnotesize Latitude};
    \node[below=of img, node distance=0cm, yshift=1.2cm,xshift = -1.2cm,font=\color{black}] {\footnotesize Longitude};
    \node[below=of img, node distance=0cm, yshift=1.2cm,xshift = 1.6cm,font=\color{black}] {\footnotesize Depth};
    \node[below=of img, node distance=0cm, yshift=1.2cm,xshift = 4.5cm,font=\color{black}] {\footnotesize Extra Marker};
    \node[below=of img, node distance=0cm, yshift=1.2cm,xshift = 7.3cm,font=\color{black}] {\footnotesize Time intervals};
    \node[left=of img, node distance=0cm, rotate=90, anchor=center,yshift=-0.8cm,xshift=0cm,font=\color{black}] {\footnotesize Earthquake};
    \end{tikzpicture}
    \end{minipage}
    \vspace{0.2cm}
    \\
\centering
    \begin{minipage}{1\linewidth}
    \begin{tikzpicture}
    \node (img)  {\includegraphics[width=1\linewidth]{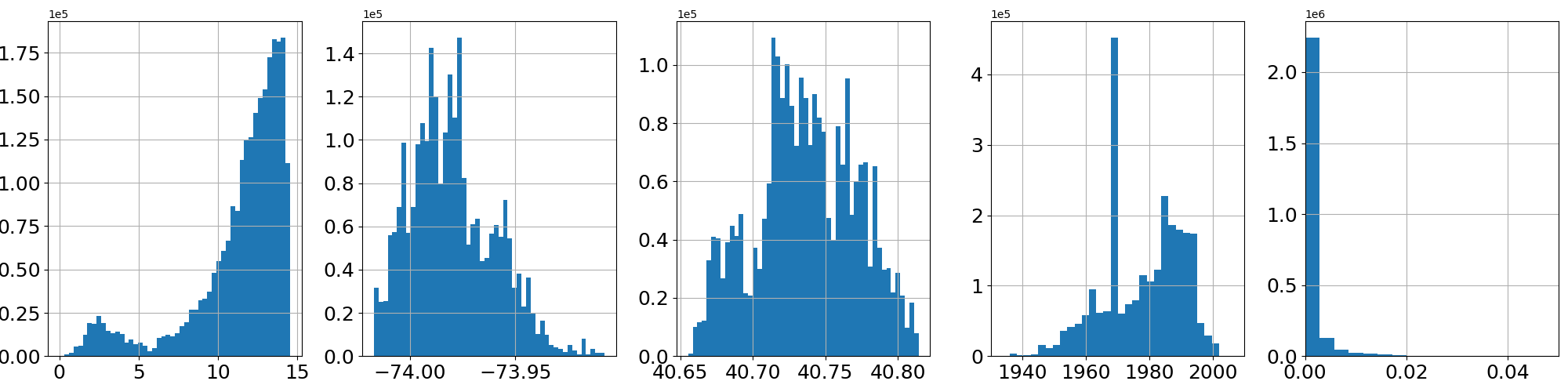}};
    \node[left=of img, node distance=0cm, rotate=90, anchor=center,yshift=-0.8cm,xshift=0cm,font=\color{black}] {\footnotesize Citibike};
    \end{tikzpicture}
    \end{minipage}
    \vspace{-0.3cm}
    \\
    \centering
    \begin{minipage}{1\linewidth}
    \begin{tikzpicture}
    \node (img)  {\includegraphics[width=1\linewidth]{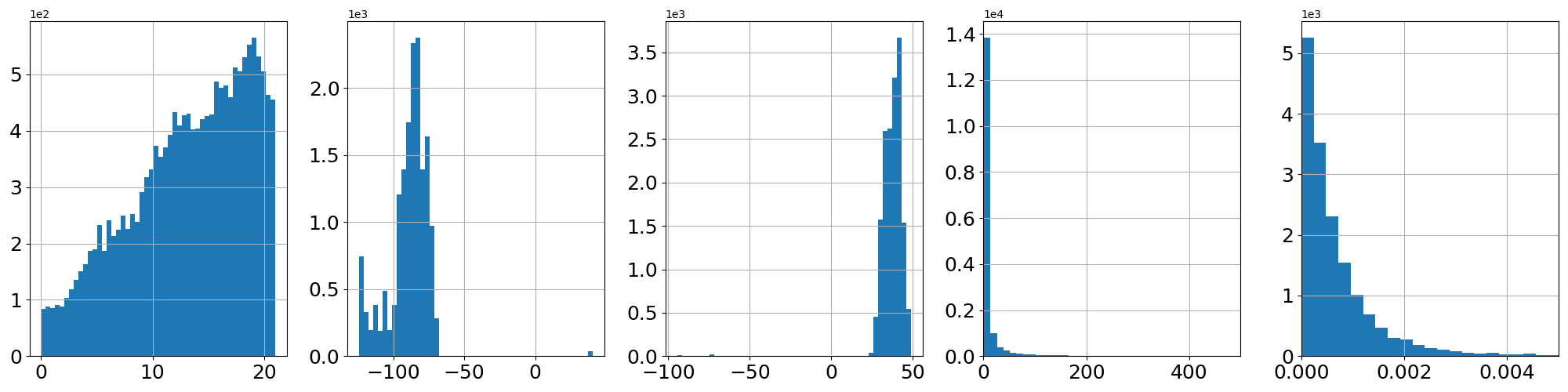}};
    \node[left=of img, node distance=0cm, rotate=90, anchor=center,yshift=-0.7cm,xshift=0cm,font=\color{black}] {\footnotesize Covid19};
\end{tikzpicture}
\end{minipage}
\vspace{-0.3cm}
\\
    \centering
    \begin{minipage}{1\linewidth}
    \begin{tikzpicture}
    \node (img)  {\includegraphics[width=1.03\linewidth]{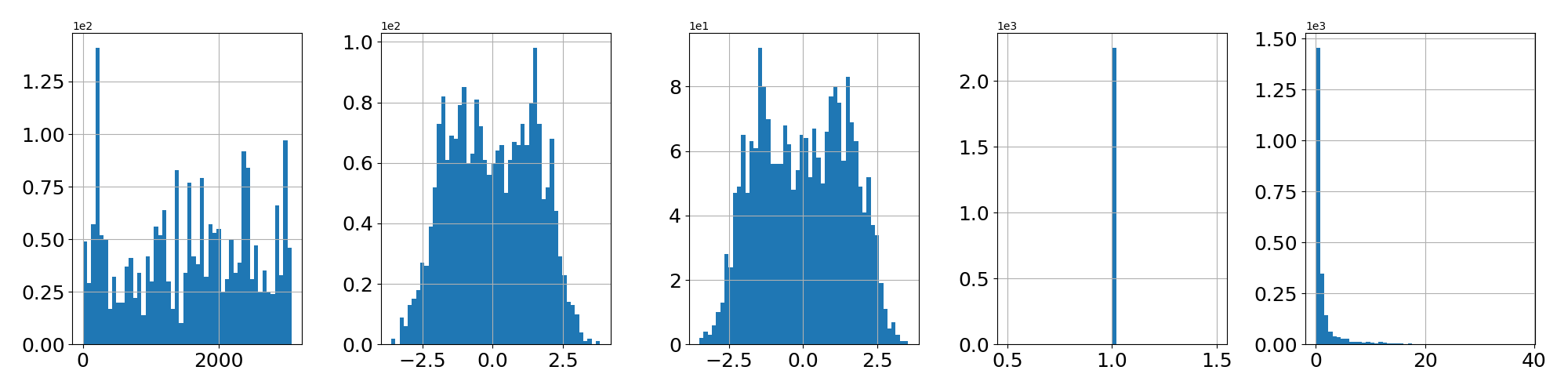}};
    \node[below=of img, node distance=0cm, yshift=1.2cm,xshift = -6.6cm,font=\color{black}] {\footnotesize Time};
    \node[below=of img, node distance=0cm, yshift=1.2cm,xshift = -3.4cm,font=\color{black}] {\footnotesize Latitude};
    \node[below=of img, node distance=0cm, yshift=1.2cm,xshift = -0.0cm,font=\color{black}] {\footnotesize Longitude};
    \node[below=of img, node distance=0cm, yshift=1.2cm,xshift = 3.5cm,font=\color{black}] {\footnotesize Extra Marker};
    \node[below=of img, node distance=0cm, yshift=1.2cm,xshift = 7.1cm,font=\color{black}] {\footnotesize Time intervals};
    \node[left=of img, node distance=0cm, rotate=90, anchor=center,yshift=-1.0cm,xshift=0cm,font=\color{black}] {\footnotesize Pinwheel};
\end{tikzpicture}
\end{minipage}
\hspace{0.4cm}
\centering
    \\
    \begin{minipage}{1\linewidth}
        \caption{Histograms of different markers associated with different datasets used in this work. The "extra marker"s associated with Earthquake, Citibike, Covid-19, and pinwheel datasets are the magnitude, birth year, number of cases, and all-one intensity, respectively. }
\label{fig:Datahistograms}
    \end{minipage}
\end{figure}

\section{Hyper-parameters}
\label{app:hyperparams}

We have set the number of attention layers in both the encoder and the decoder to 6, whereas each layer contains 6 heads individually. The reason behind these choices are what proposed by the main transformer network proposed in NLP \cite{vaswani2017}, where they suggest to choose between 6 to 8 attention heads and layers. Before inputting any of the sequences to the encoder and the decoder, we first used Multi-layer perceptron (MLP) layers with ELU activation functions, to embed to a 64 dimensional space. The latent representations $\mathbf{h}_{t_i}$,  $\mathbf{h}_{{\mathbf{x}}_i}$, depicted in Figure~\ref{fig:network} are in a 64 dimensional space, whereas we used another set of MLP layers with ELU activation functions to transfer back $h_{t_l}$, and $\mathbf{h}_{{\mathbf{x}}_l}$ to the same dimension as $t_l$, and ${\mathbf{x}}_l$ before injecting them into the probabilistic layers. This was due to the use of bijective layers (normalizing flows) following the probabilistic layers only being capable of mapping to the same dimensional space.

\begin{table}[h!]
\caption{Hyper-parameters and layer-types used in our model for both training and test.}
\label{table: hyper-params}
\begin{center}
\begin{tabular}{llll}
\multicolumn{1}{c}{\bf parameter}  &\multicolumn{1}{c}{\bf value} &\multicolumn{1}{c}{\bf parameter} &\multicolumn{1}{c}{\bf value}
\\ \hline \\
Epochs         & 1k &Batch-size & 32\\
Sequence-length             & 500 & learning-rate (starting)& 1e-3\\
Input-length             & 497 & Output-length & 3\\
Attention layers             & 6 & Attention heads& 6\\
Marker-embedding dim & 64 & Dropout-rate            & 0.1$\sim$ 0.2 \\ 
$\lambda_1$ (for $L_1$ norm)             & 0.1 & $\lambda_2$ (for $L_2$ norm) & 0.1
\\ \hline \\
\multicolumn{1}{c}{\bf probabilistic layer}  &\multicolumn{1}{c}{\bf type} &\multicolumn{1}{c}{\bf bijective layer} &\multicolumn{1}{c}{\bf type}
\\ \hline \\
time & exponential & time & softsign\\
space & multi-variate Gaussian & space & RealNVP\\
\end{tabular}
\end{center}
\end{table}

As mentioned earlier in Section~\ref{sec:network}, we used multi-variate Gaussian and exponential probabilistic layers for space and time, respectively as considered by the Hawkes model. The markers histograms shown by Figure~\ref{fig:Datahistograms} as well indicate the usefulness of these choices for our probabilistic layers. We also tried modeling the time distribution using Gaussian probabilistic layers which led to bad results. 

For the bijective flow layers we tried using RealNVP \cite{dinh2016density} and the Masked Autoregressive flows \cite{papamakarios2017masked} for the space distributions, and we tried softplus and softsign flows for time. Our results indicate that the RealNVP and softplus are capable of better modeling the expected space and time batched distributions. Note that both the RealNVP and Masked Autoregressive flows are only applicable to data with dimension of higher than 1, and therefore are not good candidates for transforming the time batched distributions.

To overcome the overfitting when training our model, we used dropout layers with rates listed as $\{0.1, 0.15, 0.2\}$. For the earthquake and Hawkes pinwheel synthetic data, both $0.1$ and $0.15$ were good options whereas for the Citibike and Covid-19 we had to increase the rate to $0.2$ to reduce the overfitting. 

All the mentioned hyperparameters can be set differently in our network using argparsing\footnote{Code for preprocessing and training are open-sourced at \url{https://github.com/Negar-Erfanian/Neural-spatio-temporal-probabilistic-transformers}}

\end{document}